\documentclass[AMA,STIX1COL]{WileyNJD-v2}

\pdfoutput=1
\usepackage{epstopdf}
\usepackage{graphicx}  %插入图片的宏包
\usepackage{float}  %设置图片浮动位置的宏包
\usepackage{subfigure}  %插入多图时用子图显示的宏包
\usepackage{makecell}
\usepackage{url}
\usepackage{booktabs}
\usepackage{rotating}
\usepackage{multirow}
\usepackage{longtable}

\articletype{Article Type}%

\received{}
\revised{}
\accepted{}

\vspace{-2 mm}

\begin{document}

%\title{This is the sample article title\protect\thanks{This is an example for title footnote.}}
\title{A Self-adaptive SAC-PID Control Approach based on Reinforcement Learning for Mobile Robots}
\author[1]{Xinyi Yu}

\author[1]{Yuehai Fan}

\author[1]{Siyu Xu}

\author[1]{Linlin Ou*}

\authormark{AUTHOR ONE \textsc{et al}}

\address[1]{\orgdiv{College of Information Engineering}, \orgname{Zhejiang University of Technology}, \orgaddress{\state{Zhejiang}, \country{China}}}

\corres{*Linlin Ou, College of Information Engineering, Zhejiang University of Technology, Hangzhou, Zhejiang, China. \email{linlinou@zjut.edu.cn}}

\presentaddress{College of Information Engineering, Zhejiang University of Technology, Hangzhou, Zhejiang, China}

\abstract[Abstract]{Proportional-integral-derivative (PID) control is the most widely used in industrial control, robot control and other fields. However, traditional PID control is not competent when the system cannot be accurately modeled and the operating environment is variable in real time. To tackle these problems, we propose a self-adaptive model-free SAC-PID control approach based on reinforcement learning for automatic control of mobile robots. A new hierarchical structure is developed, which includes the upper controller based on soft actor-critic (SAC), one of the most competitive continuous control algorithms, and the lower controller based on incremental PID controller. Soft actor-critic receives the dynamic information of the mobile robot as input, and simultaneously outputs the optimal parameters of incremental PID controllers to compensate for the error between the path and the mobile robot in real time. In addition, the combination of 24-neighborhood method and polynomial fitting is developed to improve the adaptability of SAC-PID control method to complex environments. The effectiveness of the SAC-PID control method is verified with several different difficulty paths both on  Gazebo and real mecanum mobile robot. Futhermore, compared with fuzzy PID control, the SAC-PID method has merits of strong robustness, generalization and real-time performance.}
%Proportional-integral-derivative (PID) control is the most widely used in industrial control, robot control and other fields. However, traditional PID control is not applicable when the system cannot be accurately modeled and the operating environment is variable in real-time. To tackle these problems, we propose a self-adaptive SAC-PID control approach based on reinforcement learning for automatic control of line-following robots. For the formulation of self-adaptive PID controller, based on off-policy algorithm version of the soft actor-critic(SAC), we develop a model-free SAC-PID control method for real-time self-tuning parameters of PID controllers. The proposed SAC-PID control approach uses a new hierarchical structure including the soft actor-critic layer as the upper controller and the incremental PID as the lower controller. In this paper, the upper controller tunes the parameters of the lower controller dynamically according to the current error between the center of the mobile robot and the path. After receiving the actions given by SAC, the lower controller adjusts the velocities of mobile robots to follow the line. The above method was tested in several different difficulty paths in Gazebo and ROS presenting the strong robustness, generalization and real-time performance of this approach.}


\jnlcitation{\cname{%
\author{},
\author{},
\author{},
\author{}, and
\author{}} (),
\ctitle{A regime analysis of Atlantic winter jet variability applied to evaluate HadGEM3-GC2}, \cjournal{Q.J.R. Meteorol. Soc.}, \cvol{2017;00:1--6}.}

\maketitle

%\footnotetext{\textbf{Abbreviations:} ANA, anti-nuclear antibodies; APC, antigen-presenting cells; IRF, interferon regulatory factor}

\section{Introduction}\label{sec1}
Proportional integral derivative (PID) control is the most widely used control method in industrial and robotic control because of its simple structure, strong robustness, strong adaptability and the fact that the parameter adjustment and tuning rarely depends on the specific model of the controlled object. Actually, traditional PID control often uses the experience tuning method to obtain the appropriate PID parameters. Therefore, modern control theory has contributed to many tuning methods. Hernández-Diez\cite{Jose2019Practical} proposed a parameter adjusting approach for PID controllers which partition the parameter space in to regions, defined by separating hyperplane, with constant number of unstable roots. Onat\cite{2018Onat} developed a graphical method for tuning the PI–PD controller parameters based on the stability boundary locus by equating the real and the imaginary parts of the characteristic equation to zero.

However, the above tuning methods often need accurate mathematical model and complex calculation. The PID tuning techniques tend to have a poor performance requiring an exhaustive expert knowledge about the system behavior and its variant operational environments. In this sense, the classical PID control cannot be modified in real time during the control process of the robot and make it difficult to realize the real-time control and optimal control. Yang\cite{yang2021new} designed a new adaptive motion controller for robot manipulators subject to unmeasurable or uncertain plant parameters. Besides, Yang\cite{yang2021adaptive} proposed the controller based on adaptive fuzzy control approach to solve the control problem for a class of MIMO underactuated systems with plant uncertainties and actuator deadzones. These intelligent control methods have the ability to adapt the changes of environment or the performance of robots, which seems to be suitable for solving these types of problems. Futhermore, adaptive parameter tuning control algorithm for PID controllers began to arise. For example, Papadopoulos \cite{papadopoulos2012automatic} proposed an automatic PID tuning method based on amplitude optimization principle for single-input and single-output control problem of a linear dynamic model. However, traditional adaptive PID control methods usually fail to achieve optimal control performance, especially for mobile robots operating in unknown environments, such as automatic line-following robots with unknown paths.

%With the development of industrial technology, modern engineering systems are developing in the direction of large-scale and complex.
In recent years, automatic line-following robots have been increasingly used in industrial scenes, such as material escorting robots in automatic production lines and fire-fighting inspection robots in fire-prone areas\cite{khoon2012autonomous,mae2012modified}, because they can complete some high-workload or high-risk tasks in complex and harsh applications. In addition, the line-following robot has been widely used in civil, military and medical fields\cite{punetha2013development,ilias2014hospital}. Line-following robot is a kind of robot that can move autonomously according to a given route mark which is usually represented by black line on a white surface or other color combinations. The sensors are required for automatic line-following robots to extract the external route information and transmit it to the control system. These sensors can be grayscale sensors, electromagnetic sensors and cameras. According to the obtained route information, the control system manipulates the robot to maintain the route, and continuously corrects the deviation from the route which it is following at the same time. Engin \cite{engin2012path} successfully realized the tracking of straight, circular, sharp turning and S-shaped line by using PID control algorithm, and improved the navigation accuracy. Juang \cite{juang2019visual} proposed a slope matching algorithm for the route tracking of humanoid robots. The camera was employed to capture the route information, and the curvature and slope estimation parameters of the route obtained from image processing were applied as the input of PID controller to reduce the tracking error. However, due to the problems such as unknown route and low modeling accuracy of mobile robots, it is difficult to obtain the optimal parameters of PID controllers in real time only by using traditional adaptive parameter tuning control methods.

%With the development of artificial intelligence, reinforcement learning(RL) algorithms have attracted the attention of academia and industry in recent years.
For adaptive control formulations of difficult modeling controlled objects in complex and variant environments, such as line-following robots, reinforcement learning(RL) is one of the most powerful learning paradigms.
Different from the traditional control methods, reinforcement learning \cite{sutton2018reinforcement} can not only interact with the environment autonomously and learn the optimal policy through repeated trials, which has certain robustness and real-time performance, but also solve complex problems through model-free learning algorithm \cite{8460756} for some control objects that are difficult to model, which has certain generalization. The previous reinforcement learning method \cite{gosavi2009reinforcement} is based on the optimal control theory in which the sequence decision problem is considered as an adaptive dynamic programming problem. An optimization policy algorithm based on the value function was further proposed as evaluation criterion\cite{watkins1992q}. However, traditional reinforcement learning is limited to low-dimensional and simple scenarios due to the lack of good state and policy representation ability. It is difficult to converge to an optimal policy for high-dimensional state. Mnih \cite{mnih2013playing} proposed a deep Q network model DQN that combines the advantages of feature representation in deep learning and real-time optimization policy for reinforcement learning. Since then, deep reinforcement learning has been widely used in game development \cite{berner2019dota,ye2019mastering}, natural language processing \cite{lewis2017deal,weisz2018sample}, autonomous driving \cite{ye2020automated}, recommended search systems\cite{derhami2015web}, robotic skills learning \cite{zeng2018learning,zeng2020tossingbot} and other fields.

Reinforcement learning has been incerasingly used in the field of PID control to realize the adaptive adjustment of PID parameters.
%In order to minimize the defined cost function, Boubertakh \cite{boubertakh2010tuning} used the Q-learning algorithm to adjust the fuzzy rules of PI and PD controllers online.
Carlucho \cite{carlucho2017incremental} proposed an incremental Q-learning algorithm to learn and tune the parameters of the PID controller online, and the effectiveness of this method was proved on a real mobile robot. Based on this, Carlucho \cite{Carlucho2019Double} also developed the double Q-learning algorithm for on-line autonomous adaptation of low-level PID controllers of mobile robots. The Q-learning algorithm based on value function has strong ability of policy evaluation, but it is difficult to obtain effective solutions for the problems of continuous action space or high-dimensional action space. Policy-based method is another way to optimize policy, compared with the value-based method, which has better convergence and can deal with the problems of continuous action space and stochastic policy, but it is easy to fall into local optimum. Therefore, Konda\cite{konda2000actor} proposed an actor-critic algorithm which combines the value-based method and the policy-based method. Akbarimajd \cite{akbarimajd2015reinforcement} used the adaptive PID controller based on the actor-critic algorithm and drove the two-degree-of-freedom of the robotic manipulator. Carlucho \cite{carlucho2020adaptive} presented a model free goal-driven deep RL method based on the deep deterministic policy gradient (DDPG) algorithm \cite{lillicrap2015continuous} for self-tuning of the low-level PID controllers of mobile robots. The proposed hybrid control policy adopts an actor-critic structure and a gradient inversion scheme to constrain the actor outputs along the training phase. However, because of the combined effect of the deterministic actor network and Q-function, DDPG also has some disadvantages, such as the difficulty in stabilization and sensitivity to hyperparameters. In order to solve the above problems, Haarnoja \cite{haarnoja2018soft}  proposed an off-policy maximum entropy actor-critic algorithm with a stochastic actor, that is soft actor-critic (SAC), which ensures the effective learning of samples and the system stability. The above methods of combining RL and PID controller are all applied to specified target velocity for tracking, but the tracking situation when the target velocity is constantly changing is not discussed, such as line-following task. As an alternative, Saadatrnand\cite{saadatmand2020autonomous} proposed a MIMO simulated annealing (SA) based on Q-learning method to control a line following robot. However, this method requires precise physical modeling and the application scenarios are more limited. Matczak\cite{matczak2017dim} proposed tracking algorithm based on deep learning for the line-following robot, which requires large training image and high-performance computing equipment.
%Based on SAC algorithm, Fu \cite{fu2020soft} suggested a novel live video transcoding and streaming scheme, which improved video bitrate and optimized vehicle scheduling. Barros \cite{barros2020using} presented a framework to train the SAC algorithm for low-level control of a quadrotor in a go-to-target task.

%In views of the advantages of the SAC algorithm, in this work, we propose a model-free adaptive SAC-PID method based on soft actor-critic for the line-following robot with unknown and complex routes including forks and sharp turns. The proposed SAC-PID control method uses a new hierarchical structure including soft actor-critic layer and incremental PID layer. Soft actor-critic receives the information of the mobile robot as input, and simultaneously outputs the optimal parameters of the incremental PID at the current time, which compensates for the error between the path and the mobile robot in real time. In addition, we also proposed a combination of 24-neighborhood method and polynomial fitting, which effectively extracts the correct boundary of the fork and reduces the disturbance caused by the change of environment. Finally, the strong robustness, generalization and real-time performance of the SAC-PID control approach are verified through simulations of different difficulty paths in Gazebo and ROS. The results show that a high success rate can be obtained. In our knowledge this is among, if not, we are the first to implement model-free adaptive PID based on reinforcement learning to control line-following robots.
Inspired by RL and the automatic line-following method based on PID controller, we propose a model-free adaptive SAC-PID control method based on soft actor-critic for the line-following robot with unknown and complex routes including forks and sharp turns. Compared with above method, our proposed method can adjust optimal parameters of PID controller to track the changing target velocity in real time without accurate physical models of mobile robots and large training data. The proposed SAC-PID control method uses a new hierarchical structure including soft actor-critic layer and incremental PID layer. Soft actor-critic receives the information of the mobile robot as input, and simultaneously outputs the optimal parameters of the incremental PID controllers at the current time, which compensates for the error between the path and the mobile robot in real time. In addition, we also proposed a combination of 24-neighborhood method and polynomial fitting, which effectively extracts the correct boundary of the fork and reduces the disturbance caused by the change of environment. Compared with fuzzy PID control, the strong robustness, generalization and real-time performance of the SAC-PID control approach are verified through simulations of different difficulty paths in Gazebo. Furthermore, we also demonstrate the effectiveness of the SAC-PID control method on a mecanum mobile Robot in real world environments. In our knowledge this is among, if not, we are the first to implement model-free adaptive PID based on reinforcement learning to control line-following robots. Qualitative results are available at \url{https://youtu.be/GaWI_T6etUM}.

The remainder of this paper is organized as follows: In section \ref{sec2}, the problem statement is presented. In section \ref{sec3} we indicate the details of the SAC-PID control approach for line-following robot. The simulation and physical experiment results of this SAC-PID control method presenting in section \ref{sec4}. Finally, we draw a conclusion in section \ref{sec5}.

\section{problem statement and analysis}\label{sec2}

Traditional PID control is difficult to complete the complex control task in unknown and variant environment, and the parameters of PID controllers need to be adjusted manually with expert knowledge. To deal with the complex control environment and avoid the tedious process of tuning PID parameters, a model-free adaptive SAC-PID control method will be studied in this paper to solve the automatic adjustment problem of the PID control parameters. In this section, we introduce the main elements that support our proposal.

\subsection{Incremental PID controllers in mobile robots}
For a computer-controlled line-following robot, the control algorithm should be discrete rather than continuous. Different from the typical continuous PID control law, the digital PID control law is discrete which can generally be divided into positional PID control and incremental PID control. Once the control output $u(t)$ is wrong, using positional PID controllers will cause a large change of the system because of the accumulation of all the past errors. Consequently, the positional PID control has poor robustness and heavy calculation workload, compared with the incremental PID control which is only related to the error sampling value of the last three moments. In addition, the output of incremental PID control is the increment of controller output $\vartriangle u(t)$, so the incremental PID control has less influence on the system and strong robustness compared with the positional PID control with direct output $u(t)$. The incremental PID control law is as follows:
\begin{eqnarray}
\begin{aligned}
u(t)&=u(t-1)+\vartriangle{u(t)}\\
&=u(t-1)+k_p[e(t)-e(t-1)]+k_ie(t)+k_d[e(t)-2e(t-1)+e(t-2)]
\end{aligned}
\end{eqnarray}
where $t$ is the discrete sampling time, $k_p, k_i$ and $k_d$ are the proportional, integral and differential coefficients of the PID controller respectively, $\vartriangle{u(t)}$ is the output of the incremental PID controller at the current time $t$, that is, the difference between the control output $u(t)$ at the current time $t$ and the control output $u(t-1)$ at the previous time $t-1$, and $e(t)$, $e(t-1)$ and $e(t-2)$ are the system errors of the time $t$, $t-1$ and $t-2$ respectively.

Generally, in the application of mobile robot, the dynamic performance of the robot is not completely predictable in advance and may change with the external environment, so it is difficult to establish an accurate mathematical model of the mobile robot. What's more, the route is often unknown, so it is necessary to consider the adaptive PID control. The adaptive PID control can select the optimal PID parameters in real-time according to the dynamic performance and operating environment of the mobile robot and this is where artificial intelligence can be of aid. Compared with traditional adaptive PID control methods such as fuzzy-PID, model-free self-adaptive PID controller based on reinforcement learning can adjust PID parameters better, especially when encountering complex unknown routes, such as route line crossing, sharp turning. In the next section, the RL formulation to obtain the gains of the PID controller is introduced.

\subsection{Deep reinforcement learning for PID tuning}
The basic process of reinforcement learning that is a Markov decision process (MDP) represented by a tuple:$\left\{\mathcal{S},\mathcal{A},\mathcal{P},r\right\}$, where $\mathcal{S}$ is a set of state, $\mathcal{A}$ is a set of action, $\mathcal{P}: \mathcal{S}\times\mathcal{A}\times\mathcal{S}\rightarrow\mathbb{R}$  is transition probability, $r: \mathcal{S}\times\mathcal{A}\rightarrow\mathbb{R}$ is the reward function. %Reinforcement learning process is shown in Fig. \ref{MDP}.
Agent observes the state $s_t$ at time $t$, and then selects an action $a_t$. The state transfers from $s_t$ to $s_{t+1}$ according to probability $\mathcal{P}(s_{t+1}|s_t,a_t)$. At the same time of state transitions, agent receives reward $r_t(s_t,a_t)$, where $s_t\in\mathcal{S}$, $a_t\in\mathcal{A}$,  The cumulative reward $G_t$ obtained at the end of the process is given by
\begin{eqnarray}
G_t=r_t+\gamma r_{t+1}+\gamma^2r_{t+2}+\cdot\cdot\cdot= \sum_{k=0}\gamma^kr_{k+1}
\end{eqnarray}
where $\gamma$ is discount factor.
The goal of reinforcement learning is to obtain the optimal policy $\pi^*$, that is, to maximize the cumulative reward $G_t$ while reaching the target state.

Compared with standard RL, the SAC introduced a more general maximum entropy objective, which favors stochastic policies $\pi$ by augmenting the objective with the expected of the policy over $\rho_{\pi}(s_t)$
\begin{eqnarray}
J(\pi)=argmax_\pi\sum^T_t\mathbb{E}_{(s_t,a_t)\sim\rho_\pi}\left[r(s_t,a_t)+\alpha\mathcal{H}(\pi(\cdot|s_t))\right]\label{entropy}
\end{eqnarray}
where $\alpha$ is the temperature parameter to determine the relative importance of entropy term against the reward and control the stochasticity of the optimal policy. The maximum entropy objective not only enables the agent to balance exploration and exploitation intelligently, but also enables the agent to capture multiple modes of near-optimal behavior.

SAC is derived starting from a maximum entropy variant of the soft policy iteration method based on a tabular setting, which consists of policy evaluation and policy improvement. In the policy evaluation step of soft policy iteration, the soft Q-value $Q(s_t,a_t)$ of a policy $\pi$ can be computed according to the maximum entropy objective in Eq. \ref{entropy}. Therefore, the modified Bellman backup operator $\mathcal{T}^\pi$ is given by
\begin{eqnarray}
\mathcal{T}^\pi Q(s_t,a_t)\triangleq r(s_t,a_t)+\gamma\mathbb{E}_{s_{t+1}\sim p}\left[V(s_{t+1})\right]\label{softqvalue}
\end{eqnarray}
where
\begin{eqnarray}
V(s_t)=\mathbb{E}_{a_t\sim \pi}\left[Q(s_t,a_t)-\alpha log\pi(a_t|s_t)\right]\label{softvalue}
\end{eqnarray}
is the soft state value function.
\begin{lemma}
Consider the soft bellman backup operator $\mathcal{T}^\pi$ in Eq. \ref{softqvalue} and a mapping $Q^0: \mathcal{S}\times\mathcal{A}\rightarrow\mathbb{R}$ with $\lvert\mathcal{A}\rvert<\infty$, $Q^{k+1}=\mathcal{T}^\pi Q^k$. Then the sequence $Q^k$ will converge to the soft Q-value of $\pi$ as $k\rightarrow\infty$. \label{lemma1}
\end{lemma}
Soft Q-function for policy $\pi$ can be obtained by repeatedly applying $\mathcal{T}^\pi$ as Lemma \ref{lemma1}.

In the policy improvement step, the policy is updated according to minimize the Kullback-Leibler (KL) divergence of the policy and Q-value policy. The new policy $\pi_{new}$ can be obtained as follows
\begin{eqnarray}
\pi_{new}=\mathop{argmin}_{\pi^{'} \in\Pi}D_{KL}\left(\pi^{'}(\cdot|s_t)||\frac{\mbox{exp}(\frac{1}{\alpha}Q^{\pi_{old}}(s_t,\cdot))}{Z^{\pi_{old}}(s_t)}\right)\label{softpolicy}
\end{eqnarray}
where $Z^{\pi_{old}(s_t)}$ is the partition function. Moreover, the policies are restricted to some set of policies $\Pi$ just like Gaussians in order to make it tractable.

\begin{lemma}
Let $\pi_{old}\in\Pi$ and let $\pi_{new}$ be the optimizer of the minimization problem defined in Eq. \ref{softpolicy}. Then $Q^{new}(s_t,a_t)\geq Q^\pi_{old}(s_t,a_t)$ for all $(s_t,a_t)\in\mathcal{S}\times\mathcal{A}$ with $|\mathcal{A}|<\infty$.  \label{lemma2}
\end{lemma}

According to Lemma \ref{lemma2}, the new policy has a higher value than the old policy with respected to the maximum entropy objective.
Iterate soft policy evaluation and soft policy improvement repeatedly from any $\pi\in\Pi$ converges to a policy $\pi^*$ such that $Q^{new}(s_t,a_t)\geq Q^\pi_{old}(s_t,a_t)$ for all  $\pi\in\Pi$ and $(s_t,a_t)\in\mathcal{S}\times\mathcal{A}$,$|\mathcal{A}|<\infty$.

%As the route and the position changes frequently during the movement of the line-following robot, the error of the mobile robot relative to the line will also change, so the PID parameters of the control robot need to be adjusted accordingly.
%It is easily to see that this problem can be extended directly to a Markov decision process as well, by providing the agent with information about current error. Therefore, the parameters $\boldsymbol{K}_t=(k_p,k_i,k_d)$ can be adjusted by RL agent.
%Specifically, a new hierarchy structure is required to control the line-following robot such that soft actor-critic is the upper controller and the PID controller is the lower controller. Soft actor-critic receives the error of the mobile robot as input, and simultaneously outputs the optimal parameters $\boldsymbol{K}_t$ of incremental PID controllers to compensate for the error in real-time.

In the vast majority of mobile robot applications, the performance of the mobile robot and the operating environment are not entirely predictable in advance. In order to maintain optimal robot performance, the parameters of its control module must be continuously adjusted. Therefore, the design of an adaptive PID controller that combines the real-time performance and optimality of RL with the strong stability of traditional PID control is considered. The design principle is that the RL agent adjusts the parameters of the lower level PID controller according to various states. It is obvious that this design principle can solve the problem that mobile robots cannot maintain optimal performance due to unpredictable environmental changes. Specifically, the RL agent makes a decision based on the state $s_t$ observed in real-time and outputs an action $a_t$ to adjust the parameters of the lower-level controller. In this case, the lower-level controller is the PID controller, so the actions are the parameters of the PID controller, i.e. $\boldsymbol{K}_t=a_t$. Then the PID controller controls the mobile robot to maintain its optimal performance based on the changing parameters $\boldsymbol{K}_t$ and the measurement error resulting from the interaction between the mobile robot and the environment. The mobile robot follows the path which is unknown and changeable, that is, the target velocity is changing at every moment. Therefore, the designed control method must have good real-time performance and stability. In addition, the optimality of the controller also needs to be considered because of some paths with large curvatures. Hence, a new hierarchy structure is required to control the line-following robot such that soft actor-critic is the upper controller and the PID controller is the lower controller. Soft actor-critic receives the error of the mobile robot as input and simultaneously outputs the optimal parameters $\boldsymbol{K}_t$ of incremental PID controllers to compensate for the error in real-time.
%As shown in Fig. \ref{overall_structure}, RL agent receives the error $e_t$ as an input state and outputs the PID controller parameters $\boldsymbol{K}_t$. Then the PID controller receives $\boldsymbol{K}_t$ adjusted by RL agent in real-time and sends output signals to the mobile robot's actuators to control the motion of the mobile robot. During the interaction, the RL agent updates the policy based on the change in error and the corresponding reward until it converges to the optimal policy. Furthermore, a new hierarchy structure is required to control the line-following robot such that soft actor-critic is the upper controller and the PID controller is the lower controller. Soft actor-critic receives the error of the mobile robot as input and simultaneously outputs the optimal parameters $\boldsymbol{K}_t$ of incremental PID controllers to compensate for the error in real-time.

%\begin{figure}[htb]
%\centerline{\includegraphics[scale=0.9]{adptivePID}}
%\caption{Adaptive PID controller architecture.\label{fig2}}
%\end{figure}

\section{SAC-PID control approach based on adaptive deep RL}\label{sec3}

When the mobile robot follows the line, the error between the path and the mobile robot is related to the current time and changes in real time, which satisfies the Markov property.
We can regard the line-following task as a Markov decision process: given a state $s_t$, the agent selects and executes an action $\boldsymbol{K}_t$ according to the policy $\pi(s_t)$. After changing the velocities of the mobile robot, the new state $s_{t+1}$ and corresponding reward $r_{t}(s_t,s_{t+1})$ will be obtained, which constitutes a set of tuples $(s_t,\boldsymbol{K}_t,s_{t+1},r_{t})$. The goal of reinforcement learning is to find a optimal policy $\pi^*(s_t)$ by the maximization of the expected cumulative reward $G_t$.

\begin{figure}[htb]
\centerline{\includegraphics[scale=0.45]{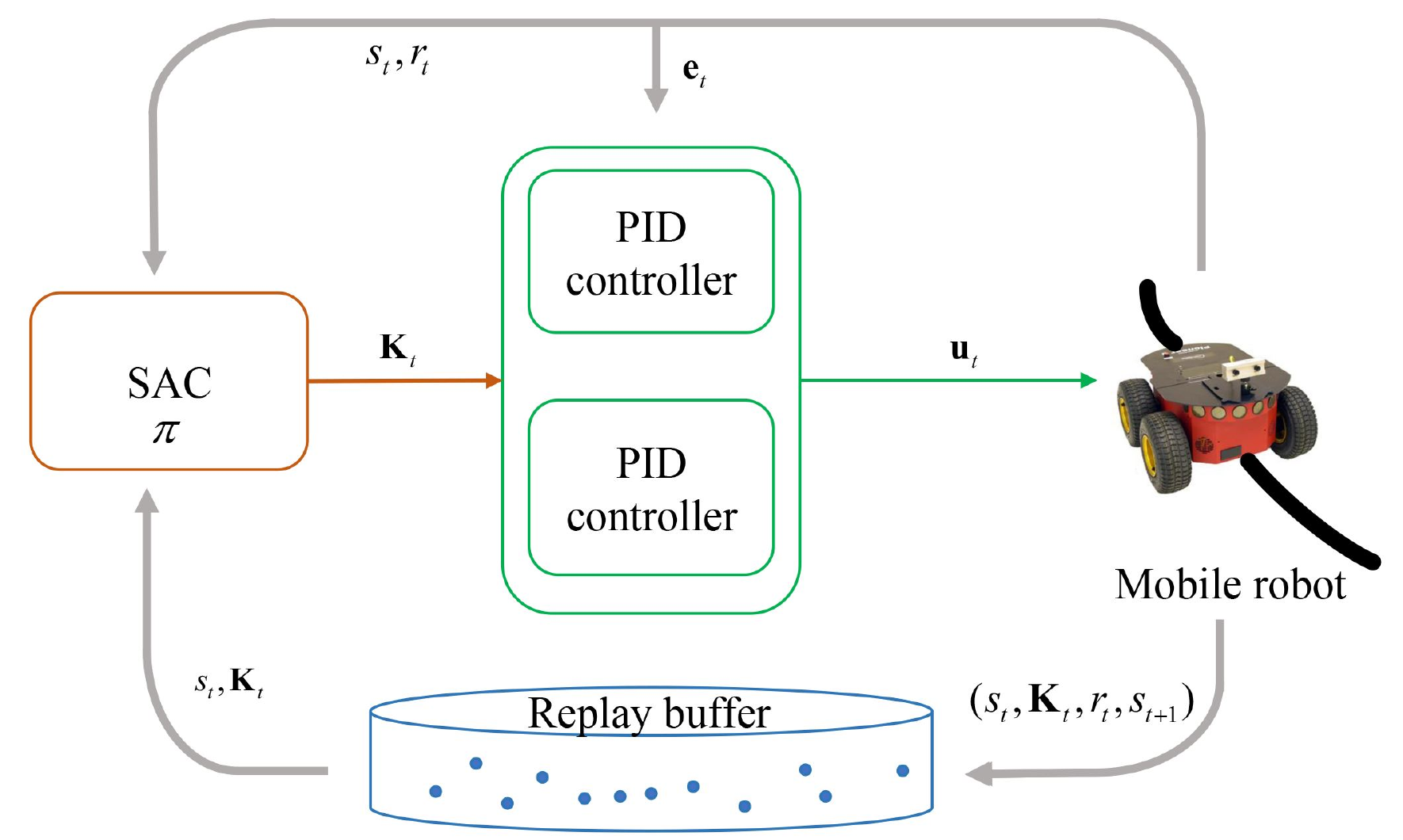}}
\caption{The overall structure of the SAC-PID control system.\label{overall_structure}}
\end{figure}

Combined with the idea of adaptive PID control, we investigate the use of off-policy maximum entropy soft actor-critic to train a stochastic policy $\pi(s_t)$ that adjust the parameters $\boldsymbol{K}_t$ of the PID controller to ensure the completion of line-following task. Based on previous soft actor-critic algorithm and incremental PID controller, a new hierarchical structure is designed for adaptive SAC-PID control approach. The overall structure is shown in Fig. \ref{overall_structure}. In this structure, the RL agent plays the role of  the upper controller, which can adjust the parameters of the lower controllers in real time according to the current state $s_t$. The lower controller compensates the error $\boldsymbol{e}_t$ by giving the output $\boldsymbol{u}_t$ to the mobile robot. In this sense, the actions $\boldsymbol{K}_t$ selected by the RL agent are the parameters of the PID controllers, such that the actor produces $\boldsymbol{K}_t=\pi{(s_t)}$ and the critic provides information for training the actor.

\subsection{The upper controller design in the SAC-PID control system}
Formally, we describe the details of the upper controller manipulating the mobile robot to complete line-following task from the aspects of policy optimization, state representations and reward function.

  In the policy optimization phase, function approximators are used for the Q-function,the value function and the policy because of the continuous line-following task. Specifically, function approximators for the Q-function, the value function and the policy are employed instead of running evaluation and improvement to convergence like soft policy iteration, alternated between optimizing both networks with stochastic gradient descent.
  The parameterized state value function $V_\psi(s_t)$ and soft Q-function $Q_\theta(s_t,\boldsymbol{K}_t)$ can be modeled as two expressive neural networks, i.e., value network and critic network.  The parameters of these networks are $\psi$ and $\theta$. Besides, a tractable policy $\pi_\phi(\boldsymbol{K}_t|s_t)$ can be modeled as a Gaussian distribution with mean and covariance given by actor neural network, where $\phi$ is the parameter of the actor network. Each of the value network, critic network and actor network, in this context, includes an input layer, two hidden layers and an output layer. The rectified linear unit (ReLU) in the hidden layer is adopted as the activation function which maps the input to the output signal. In addition, Replay buffer is constructed to record tuples $(s_t,\boldsymbol{K}_t,r_t,s_{t+1})$ and to update the parameters of value network, critic network and policy network until RL agent learns the optimal policy $\pi_\phi^*$.

Value network approximates the soft value, which can stabilize training and is convenient to train simultaneously with other networks. Value network approximator is related to critic network and policy according to Eq. \ref{softvalue}, and it can be trained to minimize the squared residual error. The objective function of value network is given by
\begin{eqnarray}
J_V(\psi)=\mathbb{E}_{s_t\sim\mathcal{D}}\left[\frac{1}{2}\left(V_\psi(s_t)-\mathbb{E}_{\boldsymbol{K}_t\sim\pi_\phi}\left[Q_\theta(s_t,\boldsymbol{K}_t)-log\pi_\phi(\boldsymbol{K}_t|s_t)\right]\right)^2\right]\label{valueupdate}
\end{eqnarray}
where $\mathcal{D}$ is the Replay buffer. Actions $\boldsymbol{K}_t$ are sampled from the current policy $\pi_\phi(\boldsymbol{K}_t|s_t)$.

The role of the critic network is to evaluate the current actor and guide the actor $\pi_\phi$ to converge to $\pi_\phi^*$, which can be trained to minimize the soft bellman residual. The objective function of critic network is presented as
\begin{eqnarray}
J_Q(\theta_i)=\mathbb{E}_{(s_t,\boldsymbol{K}_t)\sim\mathcal{D}}\left[\frac{1}{2}\left(Q_{\theta_i}(s_t,\boldsymbol{K}_t)-\left(r(s_t,\boldsymbol{K}_t)+\gamma\mathbb{E}_{s_{t+1}\sim p}\left[V_{\overline{\psi}}(s_{t+1})\right]\right)\right)^2\right]\label{criticupdate}
\end{eqnarray}
where $\overline{\psi}$ is the exponentially moving average of the value network weights\cite{mnih2015human}. In particular, two Q-functions are parameterized, with parameters $\theta_i$ to be trained independently to optimize $J_Q(\theta_i)$, where $i\in\left\{1,2\right\}$. In addition, the minimum of the Q-functions is served for value gradient and policy gradient, as proposed by Fujimoto\cite{fujimoto2018addressing}.

The actor network, which represents a Gaussian distribution, can be learned by directly minimizing the expected KL-divergence in Eq. \ref{softpolicy}. Because the sampling actions from the distribution is discrete, and it is impossible to derive and backpropagate to update the network parameters, the reparameterization trick \cite{doersch2016tutorial} is adopted to obtain actions $\boldsymbol{K}_t$:
\begin{eqnarray}
\boldsymbol{K}_t=f_\phi(\epsilon;s_t)=f_\phi^\mu(s_t)+\epsilon_t\odot f_\phi^\delta(s_t)
\end{eqnarray}
where $\epsilon_t$ is a noise vector which is sampled from some fixed distribution, in this context, normal distribution is used. $f_\phi^\mu(s_t)$ is the mean of Gaussian distribution, and $f_\phi^\delta(s_t)$ is the variance of Gaussian distribution.

Thus the objective function can be rewritten:
\begin{eqnarray}
J_\pi(\phi)=\mathbb{E}_{s_t\sim\mathcal{D},\epsilon\sim\mathcal{N}}\left[log\pi_\phi\left(f_\phi(\epsilon_t;s_t)|s_t\right)-Q_\theta\left(s_t,f_\phi(\epsilon_t;s_t)\right)\right]\label{actorupdate}
\end{eqnarray}

%state representation
For state representations, each state $s_t$ is modeled as normalized representation the coordinates of five pixels, curvature error and the velocities of the mobile robot of the scene at time $t$. The process of calculating the normalized coordinates of five pixels includes image preprocessing, boundary point extraction and centerline calculation.

\begin{figure}[t] %这里使用的是强制位置，除非真的放不下，不然就是写在哪里图就放在哪里，不会乱动
	\centering  %图片全局居中
	\vspace{-0.35cm} %设置与上面正文的距离
	\subfigtopskip=2pt %设置子图与上面正文或别的内容的距离
	\subfigbottomskip=3pt %设置第二行子图与第一行子图的距离，即下面的头与上面的脚的距离
	\subfigcapskip=-5pt %设置子图与子标题之间的距离
	\subfigure[Five pixels of black line]{
		\label{fig4}
		\includegraphics[scale=0.6]{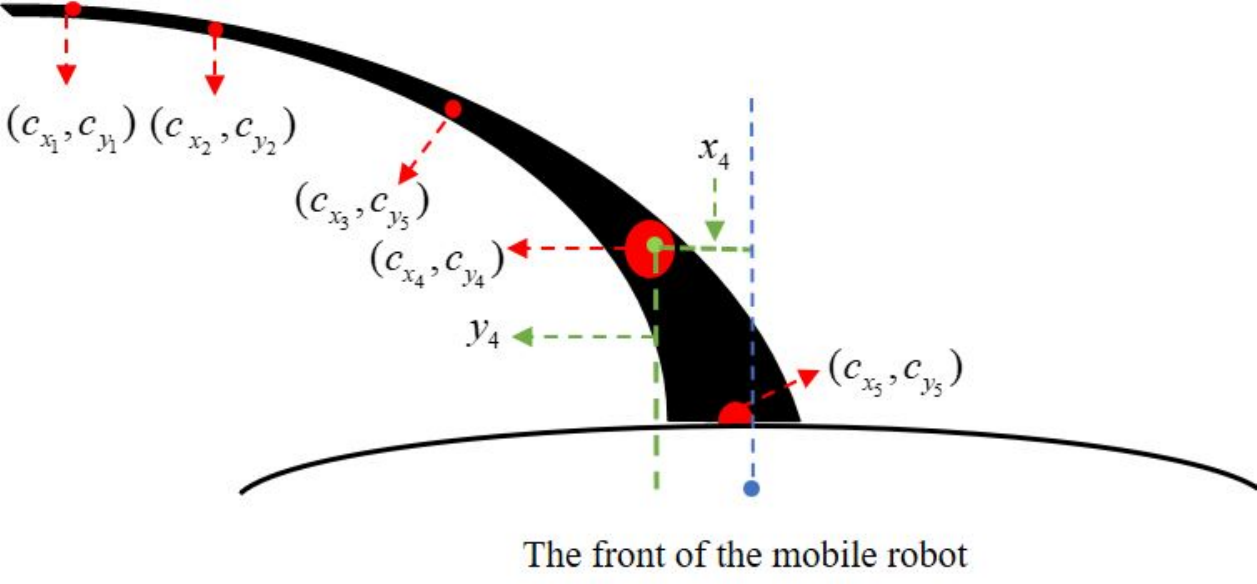}}
	\qquad %默认情况下两个子图之间空的较少，使用这个命令加大宽度
	\subfigure[The fork pixels of intersection]{
		\label{fig5}
		\includegraphics[scale=0.6]{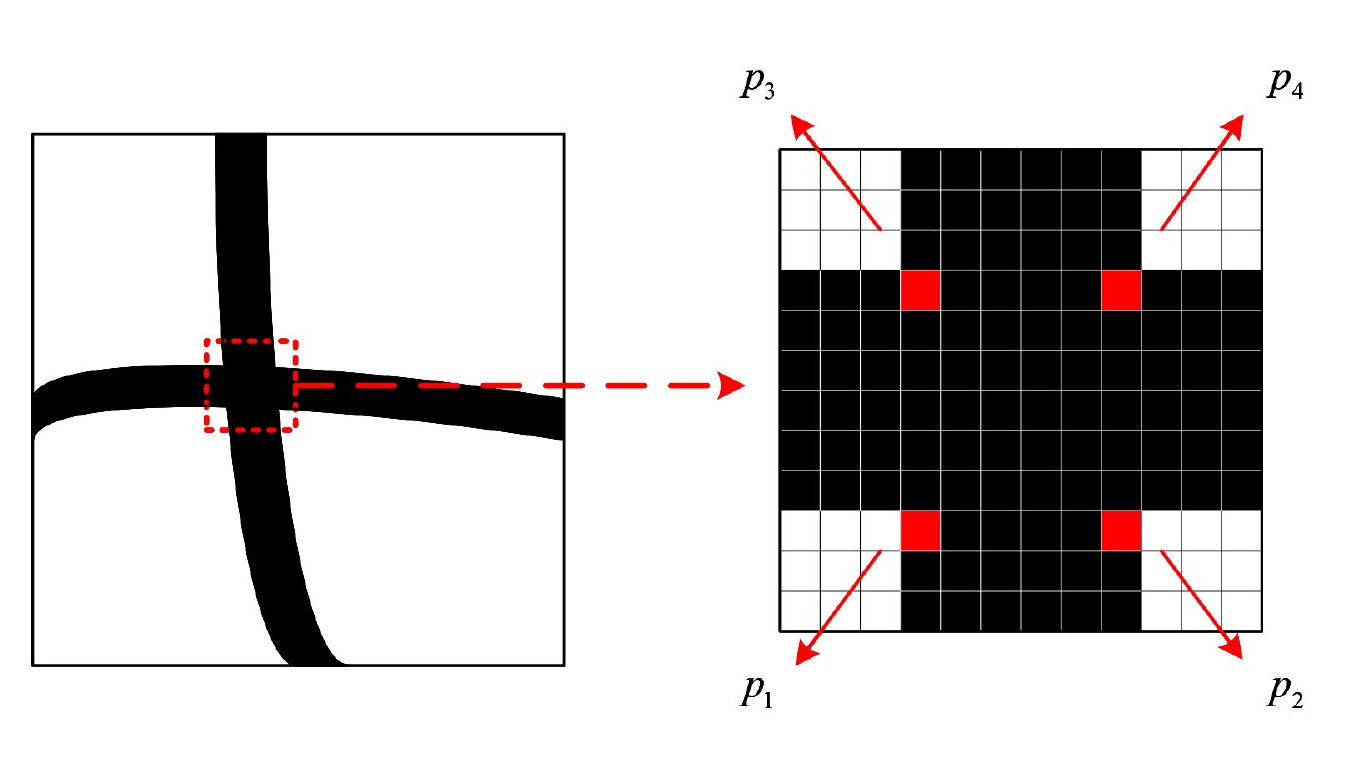}}
	  %这里是空了一行，能够实现强制将四张图分成两行两列显示，而不是放不下图了再换行，使用\\也行。
	\caption{Path 4 and corresponding performance parameters curves}
	\label{The performance of Map4}
\end{figure}

Firstly, we capture RGB image from fixed-mount camera, resize the raw image to $128\times72$ and then convert it to a binary image. Then the region growing method is used to extract all coordinates of black line boundaries in the binary image. And all the pixel coordinates of the black line are obtained. Specifically, the five pixels are located at the minimum, $1/4$, $2/4$ ,$3/4$ and the maximum vertical coordinate values of black line $(c_{x_1},c_{y_1})$,$(c_{x_2},c_{y_2})$, $(c_{x_3},c_{y_3})$, $(c_{x_4},c_{y_4})$ and $(c_{x_5},c_{y_5})$ as shown in Fig. \ref{fig4}. Finally, the five pixel coordinate values are normalized to $\left[-1,1\right]$ as the first 10 dimensions of the state $s_t$ given by
\begin{eqnarray}
[x_1,y_1,x_2,y_2,x_3,y_3,x_4,y_4,x_5,y_5]
\end{eqnarray}

The value $x_d$ is the deviation value between the black line and the front of the mobile robot at the current moment, where $d\in \{1,\cdots, 5\}$. For instance, when the point $(c_{x_4},c_{y_4})$ is on the same longitudinal horizontal line as the center point of the front of the mobile robot, ${{x}_{4}}=0$. The value $y_d$ represents the distance between its corresponding deviation value $x_d$ and the front of the mobile robot at the current moment, which is convenient for the agent to determine the importance of $x_d$ according to global information during training.

%\begin{figure}[t]
%\centerline{\includegraphics[scale=0.7]{fivepoints}}
%\caption{Five pixels of black line.\label{fig4}}
%\end{figure}
However, the mobile robot may encounter the black line crossing in the process of following the line. The traditional region growing algorithm extracts the pixel coordinates of all the boundaries of forks, which causes the black line to deviate from the true value and the mobile robot will turn quickly and lose its stability. In this work, 24-neighborhood threshold method is used to determine whether the boundary pixels are fork pixels. As shown in Fig. \ref{fig5}, there are four boundary fork pixels in an intersection, which are $p_1$, $p_2$, $p_3$ and $p_4$.
%\begin{figure}[t]
%\centerline{\includegraphics[scale=0.7]{fork}}s
%\caption{The fork pixels of intersection.\label{fig5}}
%\end{figure}

There are more black pixels in the 24-neighborhood with the fork pixel as the center than the ordinary boundary pixels. Therefore, when extracting the boundary pixels, it is necessary to calculate the sum of all pixel values in the 24-neighborhood of each boundary pixel separately to determine whether it is a fork pixel. The rule of judgment is as follows:
\begin{eqnarray}
\begin{cases}
\tau_{1}\leq \tau_{mn} \leq \tau_2     &  \mbox{if $p_{mn}$ is a fork pixel}\\
\tau_{mn} >  \tau_2      &\mbox{if $p_{mn}$ is a normal boundary pixel}
\end{cases}
\end{eqnarray}
where $\tau_{mn}=\sum\limits_{k=-2}^{k=2}\sum\limits_{j=-2}^{j=2}\tau_{(c_{x_{m+k}},c_{y_{n+j}})}$, $m\in\left\{0,\cdot\cdot\cdot,71\right\}$, $n\in\left\{0,\cdot\cdot\cdot,127\right\}$, $\tau_1$ and $\tau_2$ are the lower and upper thresholds of the sum of all pixel values in the 24-neighborhood of fork pixel.

Since the region grows from the bottom of the binary image, only $p_1$ and $p_2$ can be judged first. If it is the fork pixel $p_1$ or $p_2$, the polynomial fits the left and right boundaries to grow to $p_3$ and $p_4$, and then use $p_3$ and $p_4$ as the starting point for region growth so as to extract the complete left and right boundaries of the black line.

In order to make the mobile robot more stable at the curve line, the curvature error $e_c$ between the center line of the black line and the mobile robot itself is also added as one of the dimensions of the state. To compute this error $e_c$, we transform the projection of the RGB image to the top view, resize the image to $72\times72$, generate binary image, use region growth and 24-neighborhood threshold method to find the center line of the black line, and finally find the curvature of the center line. Thus, $e_c$ can be calculated is given by
\begin{eqnarray}
e_c=c_r-c_l
\end{eqnarray}
where $c_l$ is the curvature of the center line of the black line, $c_r$ is the curvature of the mobile robot itself.
In addition, considering that the mobile robot changes its velocity with time delay, the linear velocity $v_x$ and angular velocity $v_\omega$ of the mobile robot at time t are also part of the state.
Thus the entire $s_t$ can be expressed as an 13-dimensional vector.

\begin{eqnarray}
s_t=[x_1,y_1,x_2,y_2,x_3,y_3,x_4,y_4,x_5,y_5,e_c,v_x,v_\omega]^T
\end{eqnarray}

The idea of reward function is designed to incorporate key objectives of this study, which is to develop an automated line-following policy centered around stability and efficiency. More specifically, these evaluation indicators are explained as follows:

1) Stability: The mobile robot must always be on line, not out of line.

2) Efficiency: Evaluation of the velocity at which the mobile robot completes a full circle to complete the line-following task.

The reward function $r(t)$ is divided into three items corresponding on the three scenarios where the mobile robot reaches the goal, exceeds the boundary of the line and maintains on the line but not reaches the goal. Specifically, $r(t)$ is defined as
\[r(t)=\left\{ \begin{array}{*{35}{l}}
   {{\zeta }_{s}}s(w)+{{\zeta }_{v}}v(w)+g,&\text{                                 if  reaches the goal              }  \\
   {{\zeta }_{s}}s(w)+{{\zeta }_{v}}v(w)-g,&\text{                                 if  goes out the line              }  \\
   {{\zeta }_{r}}\frac{1}{1+{{\beta }_{1}}e(t)+{{\beta }_{2}}e(t-1)+{{\beta }_{3}}e(t-2)}&\text{     otherwise}  \\
\end{array} \right.\]
where $s(w)$ and $v(w)$ are the distance and average velocity of mobile robot in the $w-th$ episode. $e(t)$, $e(t-1)$, $e(t-2)$ are the error $x_4$ at time $t$, $t-1$ and $t-2$; and $\beta_1$, $\beta_2$ and $\beta_3$ are the corresponding weights for error $x_4$ at time $t$, $t-1$ and $t-2$. $g$ is the punishment item. $\zeta_r$, $\zeta_s$ and $\zeta_v$ are the weights of three items.
The stability of the mobile robot is related to the error at each moment. The third item of $r(t)$ is introduced to ensure the center of the mobile robot exactly on the line. Specifically, the closer the center of the mobile robot is to the center of the line, the greater the reward, and vice versa.
In terms of efficiency of evaluation indicators, the line-following robot should manage to move to the goal as soon as possible. So when the mobile robot goes out the line or reaches the goal, we will evaluate its average linear velocity as an efficiency indicator. Therefore, the physical quantity of average linear velocity is added to encourage  the mobile robot to complete the line-following task at the fastest velocity while ensuring good error correction. In addition, we also added a penalty item to encourage the mobile robot to complete the task. When the mobile robot reaches the goal, we give it positive feedback $+g$, and if it fails in the route, we give it a negative feedback $-g$.

\subsection{The lower controller design in the SAC-PID control system}

The function of the lower controller is to receive the output of actor network $\boldsymbol{K}_t$ in real time, and control the angular velocity of the mobile robot according to the control law of the incremental PID controllers. In order to make the mobile robot more stable when following the line, two parallel PID controllers, namely the main controller and the auxiliary controller, are used to jointly control the angular velocity of the robot. The control law of the main controller is indicated as below
\begin{eqnarray}
\vartriangle\omega_m=k_{mp}[e_m(t)-e_m(t-1)]+k_{mi}e_m(t)+k_{md}[e_m(t)-2e_m(t-1)+e_m(t-2)]
\end{eqnarray}
where $e_m(t)$, $e_m(t-1)$ and $e_m(t-2)$ can be considered respectively as the $x_4$ value of the $s_t$, $s_{t-1}$ and $s_{t-2}$, $\vartriangle\omega_m$ is the output of main controller and $k_{mp}$, $k_{mi}$ and $k_{md}$ are the coefficients of the first three dimensions of actor network $\boldsymbol{K}_t$. The principles for choosing the error ${{e}_{m}}$ are as follows

(1)	There exits the phenomenon that controlling the mobile robot by the computing device will cause communication lag.\label{principle 1}

(2)	The controller needs to be forward-looking to make the mobile robot follows the path with large curvature accurately. \label{priciple 2}

(3)	The center of the mobile robot is encouraged to be at the center of the path as much as possible. \label{principle 3}

${{x}_{5}}$ is close to the front of the mobile robot, which cannot meet the communication lag of principle (1) and the perspectiveness of principle (2). Therefore, ${{x}_{5}}$ cannot be selected as the error ${{e}_{m}}$.
Among ${{x}_{1}},{{x}_{2}},{{x}_{3}},{{x}_{4}}$ these four points, ${{x}_{4}}$ is not only meets the communication lag of principle (1) and the perspectiveness of principle (2), but also meets the principle (3). Therefore, ${{x}_{4}}$ is chosen for the error ${{e}_{m}}$ instead of ${{x}_{1}},{{x}_{2}},{{x}_{3}}$. The control law of the auxiliary controller is as follows

%The reason for choosing ${{e}_{m}}$ should consider the physical factors of mobile robots, such as the linear velocity and the communication lag. If ${{x}_{5}}$ is selected as the error, the mobile robot is unable to compensate for the error in time even goes out the line due to the communication lag of the mobile robot. Besides if ${{x}_{1}},{{x}_{2}},{{x}_{3}}$ are chosen as the error, it will cause the mobile robot to prejudge prematurely and go out of the line. Considering the above two situations, ${{x}_{4}}$ is chosen as the ${{e}_{m}}$ in this work.
\begin{eqnarray}
\vartriangle\omega_c=k_{cp}[e_c(t)-e_c(t-1)]+k_{ci}e_c(t)+k_{cd}[e_c(t)-2e_c(t-1)+e_c(t-2)]
\end{eqnarray}
where $e_c(t)$, $e_c(t-1)$ and $e_c(t-2)$ can be considered respectively as the $e_c$ value of $s_t$, $s_{t-1}$, $s_{t-2}$,  $\vartriangle\omega_m$ is the output of  auxiliary controller and $k_{cp}$, $k_{ci}$, $k_{cd}$ are the coefficient of the last three dimensions of actor network $\boldsymbol{K}_t$.

Thus the angular velocity of mobile robot is as follow
\begin{eqnarray}
\omega_{t}=\omega_{t-1}+\vartriangle\omega_m+\eta\vartriangle\omega_c\label{angular}
\end{eqnarray}
where $\omega_{t}$ is the angular velocity sent to the mobile robot at time $t$, $\omega_{t-1}$ is the angular velocity of the mobile robot at the time $t-1$, and $\eta$ is the proportional coefficient.

$\boldsymbol{K}_t$ can be reconstructed as follows
\begin{eqnarray}
\boldsymbol{K}_t=[k_{mp}, k_{mi}, k_{md}, k_{cp}, k_{ci}, k_{cd} ]
\end{eqnarray}

%primitive action
In the line-following task, the mobile robot has two degrees of freedom, namely linear velocity and angular velocity. In order to make the mobile robot more stable when following the line, the two-degree-of-freedom mobile robot can be decoupled, and the SAC-PID control system controls the angular velocity in real time according to the current state as shown in Eq. \ref{angular}. In order to ensure the stability of the mobile robot, a slope function can be constructed as below so that the linear velocity of the mobile robot changes with the change of error $x_4$
\begin{eqnarray}
v_x=-a_x|x_4|+b_x\label{linear velocity}
\end{eqnarray}
where $a_x$ and $b_x$ are the coefficients that control the linear velocity range of the mobile robot. When $|x_4|=1$, the linear velocity of the mobile robot changes to the minimum value. Thus the mobile robot can slowly pass a sharp turn or stabilize in time when the vibration occurs. When $|x_4|=0$, the linear velocity of the mobile robot changes to the maximum value. Hence the mobile robot can quickly follow the line when it is stable, saving time and cost.

\begin{algorithm}[htb]
  \caption{Pseudocode for the proposed algorithm SAC-PID control approach}\label{alg1}
  \begin{algorithmic}[1]
      \State Initialize parameter $\phi,\, \psi,\, \overline{\psi},\, \theta_i$ of actor network, value network and critic network, $\mbox{for}\ i\in\left\{1,2\right\}$
      \State Set up an empty Replay buffer $R$
      \For {epsiode=1 to M}
      \State Receive initial observation state $s_1$;
        \For {step=1 to N}
        \State Select actions $\boldsymbol{K}_t$ based on current actor network $\pi_\phi$
        \State  Compute angular velocity $v_\omega$ according to $s_t$ and action  $\boldsymbol{K}_t$
        \State  Compute linear velocity $v_x$ according to $s_t$
        \State  Apply velocities $(v_x,v_\omega)$ to  environment and observe next state $s_{t+1}$
        \State  Compute reward $r_t$
        \State Store following transitions $(s_t,\boldsymbol{K}_t,r_t,s_{t+1})$ into replay buffer $R$
        \If {$|R|$>b}
        \State Sample a random minibatch of $b$ transitions
        \State Update value network parameters $\psi\leftarrow\psi-\lambda_V\hat{\nabla}_\psi J_V(\psi)$
        \State Update Q-function parameters $\theta_i\leftarrow\theta_i-\lambda_Q \hat{\nabla}_{\theta_{i}}J_Q(\theta_i)$  $\mbox{for}\, i\in\left\{1,2\right\}$
        \State Update actor network parameters $\phi\leftarrow\phi-\lambda_\pi\hat{\nabla}_{\phi}J_\pi(\phi)$
        \State Update value network weights $\overline{\psi}\leftarrow\chi\psi+(1-\chi)\overline{\psi}$
        \EndIf
        \State Set $s_{t}=s_{t+1}$
      \EndFor
     \EndFor
  \end{algorithmic}
\end{algorithm}

\subsection{Algorithm Statement}
The entire algorithm process is divided into three stages, namely initialization, interaction and  optimization. In the initialization phase of the algorithm,  the parameters of the critic network $Q_\theta(s_t,a_t)$, value network $V_\psi(s_t)$ and actor network $\pi_\phi(\boldsymbol{K}_t|s_t)$ are initialized. At this stage, the previous parameter files of actor network and critic network can be loaded to continue training. In addition, an empty replay buffer to store tuple $(s_t,\boldsymbol{K}_t,r_t,s_{t+1})$ can be initialized.

For interaction phase, each episode contains a series of steps, agent interacts with environment from the first step to the last step and then enters into the next episode. Once the agent enters a new episode, it needs to observe the state as the initial state $s_1$. Then the agent samples an action $\boldsymbol{K}_t$ from current actor network according to the state $s_t$. The angular velocity $v_\omega$ and linear velocity $v_x$ of mobile robot can be calculated according to action and state $s_t$ in Eqs. \ref{angular} and \ref{linear velocity}. After it finished changing these velocities, agent will observe the current state as $s_{t+1}$.  Then the dynamic transition $(s_t,\boldsymbol{K}_t,r_t,s_{t+1})$ is stored in the replay buffer R, which will be used to train the agent.

The optimization phase is the core of the entire algorithm, which updates the network parameters.
In the training process, agent randomly samples $b$  transitions of the replay buffer to form a minibatch, which can be used for the network update. Firstly, the parameters of the value network can be updated by using the minimum of the Q-functions for value gradient in Eq. \ref{valueupdate}. Next, two Q-functions are trained independently to optimize $J_Q(\theta_i)$. Thirdly, the parameters of the actor network can be updated by using the minimum of Q-functions for policy gradient in Eq. \ref{actorupdate}. At last, $\overline{\psi}$ is updated according to exponentially moving average of the current value network weights. $\lambda_V$, $\lambda_Q$ and $\lambda_\pi$ are the learning rates, and $\chi$ is the target smoothing coefficient. Algorithm ends once all the training episodes have been accomplished. The output of the whole algorithm is the critic network and the actor network obtained after training, and the replay buffer can also be used as the output if necessary.

\begin{figure}[htp]
\centerline{\includegraphics[scale=0.4]{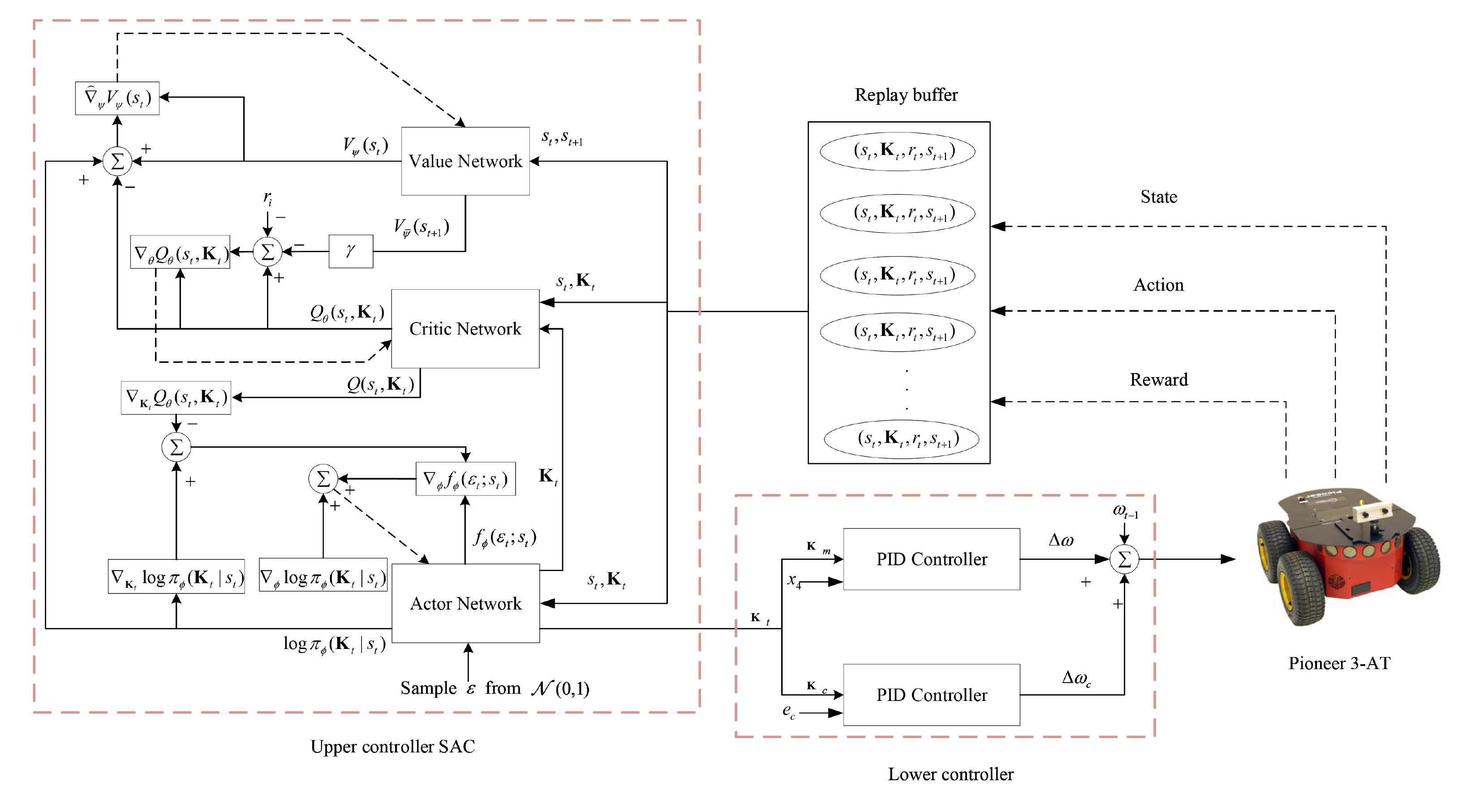}}
\caption{The specific structure of the SAC-PID control system.\label{specific_strucutre}}
\end{figure}

In order to facilitate readers understanding, the pseudo-code of the proposed SAC-PID control approach is outlined in Algorithm \ref{alg1} and the more specific structure is given in Fig. \ref{specific_strucutre}.

\section{EXPERIMENTAL RESULTS}\label{sec4}
%In order to demonstrate the robustness and real-time performance of the proposed SAC-PID control method, we test the proposed method on four different difficulty paths including forks and sharp turns. As for generalization, we randomly choose one model trained on corresponding path to test other paths. All the training and testing process are performed in ROS and Gazebo simulation environment.

In order to verify the robustness and real-time performance of the SAC-PID control method, we trained the policy of the SAC-PID control and tested it on Gazebo simulation platforms.  As for generalization, we randomly choose one model trained on the corresponding path to test other paths. Furthermore, the effectiveness of the method was further verified in hardware experiments.

\subsection{Simulation setup}
Before presenting the results of the experiment, some common aspects of all trials are introduced. Firstly the hyperparameters of SAC-PID we employed in the line-following task are listed in Appendix \ref{appendix A}. Secondly, the coefficients of reward function must provide the agent with enough information and can be implemented in different scenes. Considering that there are time delay of communication when the velocity of the line-following robot changes, the error coefficients $\beta_1$, $\beta_2$ and $\beta_3$ can be set as 0.7, 0.2 and 0.1. In terms of the overall evaluation, stability is put in the first place, followed by distance traveled and finally average velocity. Thus the overall evaluation coefficients $\zeta_r$, $\zeta_s$ and $\zeta_v$ are set as 0.5, 0.3 and 0.2. In addition, the penalty score $g$ is set to be $20$. Finally, for the threshold of the fork pixel, $\tau_1$ and $\tau_2$ are set to be 41 and 82.

\begin{figure}[t]
\centerline{\includegraphics[scale=0.15]{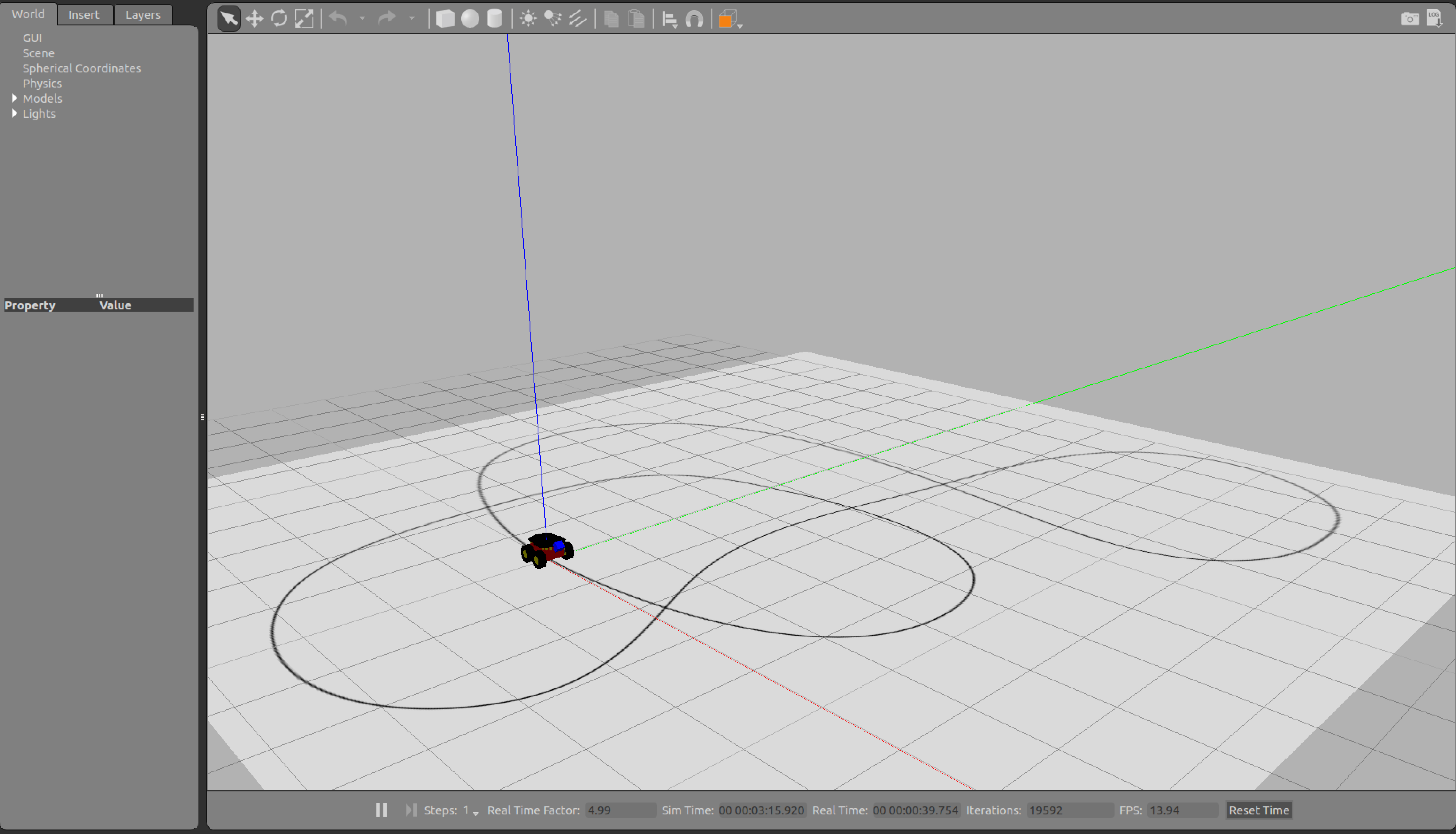}}
\caption{The simulation platform built by Gazebo and ROS.\label{The simulation platform built by Gazebo and ROS}}
\end{figure}

The training process and final test results are implemented by using Pioneer 3-AT on Gazebo simulation platform which is shown in Fig. \ref{The simulation platform built by Gazebo and ROS}. Pioneer 3-AT has 2 degrees of freedom in total, namely angular velocity $v_w$ and linear velocity $v_x$. A camera is installed in the front of Pioneer 3-AT as a sensor to collect route information. Specifically, angular velocity $v_w$ and linear velocity $v_x$ of the mobile robot can be calculated in Eqs. \ref{angular} and \ref{linear velocity}. In this context, $a_x$ and $b_x$ are set to be $0.25$ and $0.35$. Thus the range of linear velocity is $\left[0.1,0.35\right]$.

%\begin{figure}[htb] %这里使用的是强制位置，除非真的放不下，不然就是写在哪里图就放在哪里，不会乱动
	%\centering  %图片全局居中
	%\vspace{-0.35cm} %设置与上面正文的距离
	%\subfigtopskip=2pt %设置子图与上面正文或别的内容的距离
	%\subfigbottomskip=3pt %设置第二行子图与第一行子图的距离，即下面的头与上面的脚的距离
	%\subfigcapskip=-5pt %设置子图与子标题之间的距离
	%\subfigure[The side-view of the simulation platform]{
		%\label{level.sub.1}
		%\includegraphics[width=10cm]{platform_side_view}}
	%\quad %默认情况下两个子图之间空的较少，使用这个命令加大宽度
	%\qquad %默认情况下两个子图之间空的较少，使用这个命令加大宽度
	%\subfigure[The top-view of the simulation platform ]{
		%\label{level.sub.2}
		%\includegraphics[width=7cm]{platform_top_view}}
	%\caption{The simulation platform built by Gazebo and ROS}
	%\label{The platform built by Gazebo and ROS}
%\end{figure}

\subsection{Simulation experiments}

\begin{figure}[H] %这里使用的是强制位置，除非真的放不下，不然就是写在哪里图就放在哪里，不会乱动
	\centering  %图片全局居中
	\vspace{-0.35cm} %设置与上面正文的距离
	\subfigtopskip=2pt %设置子图与上面正文或别的内容的距离
	\subfigbottomskip=3pt %设置第二行子图与第一行子图的距离，即下面的头与上面的脚的距离
	\subfigcapskip=-5pt %设置子图与子标题之间的距离
	\subfigure[Path 1 and trajectory]{
		\label{map1_track}
		\includegraphics[scale=0.09]{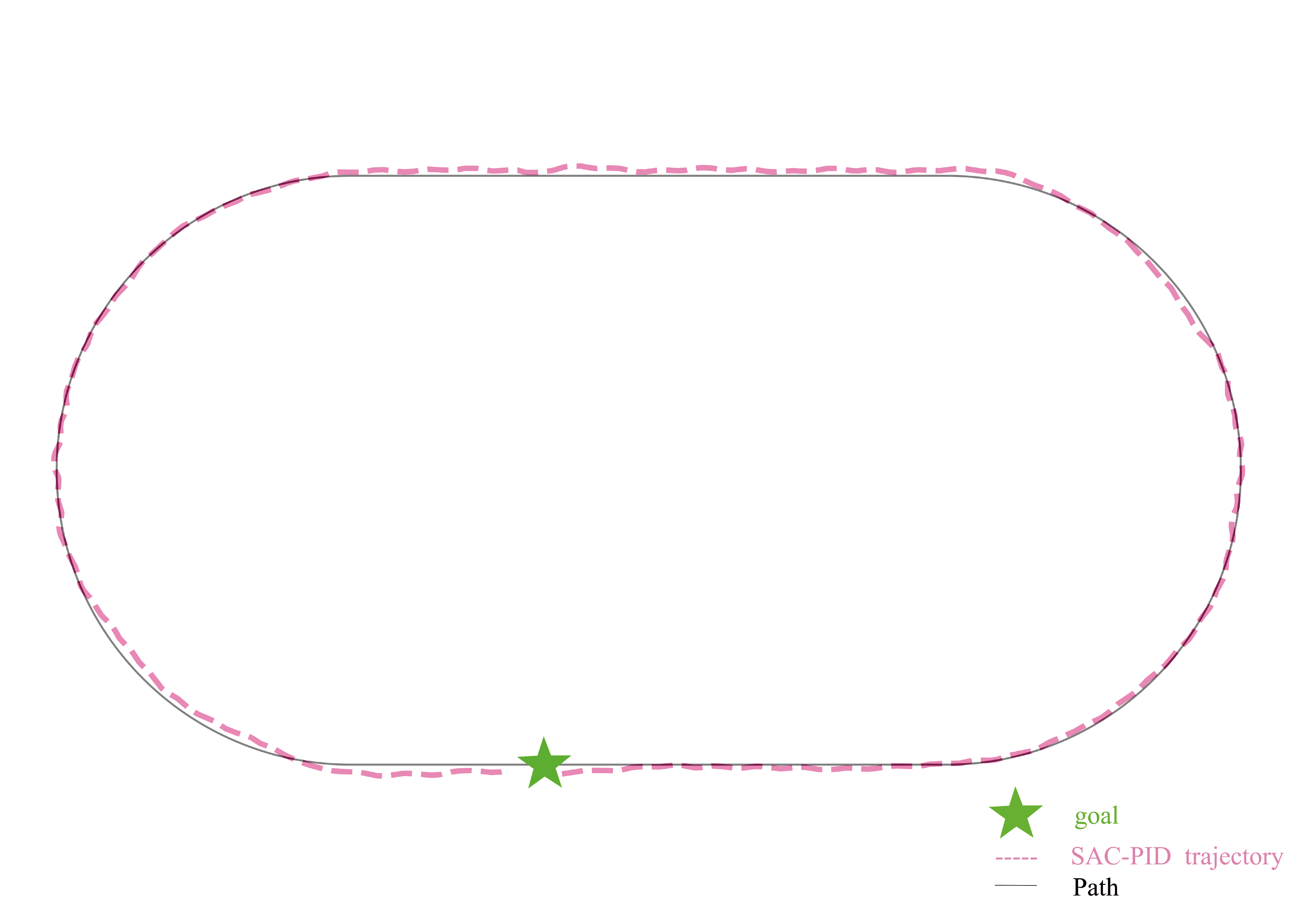}}
	\quad %默认情况下两个子图之间空的较少，使用这个命令加大宽度
	\qquad %默认情况下两个子图之间空的较少，使用这个命令加大宽度
    \quad
    \qquad
	\subfigure[Success rate]{
		\label{map1_11_24success_rate}
		\includegraphics[width=6cm]{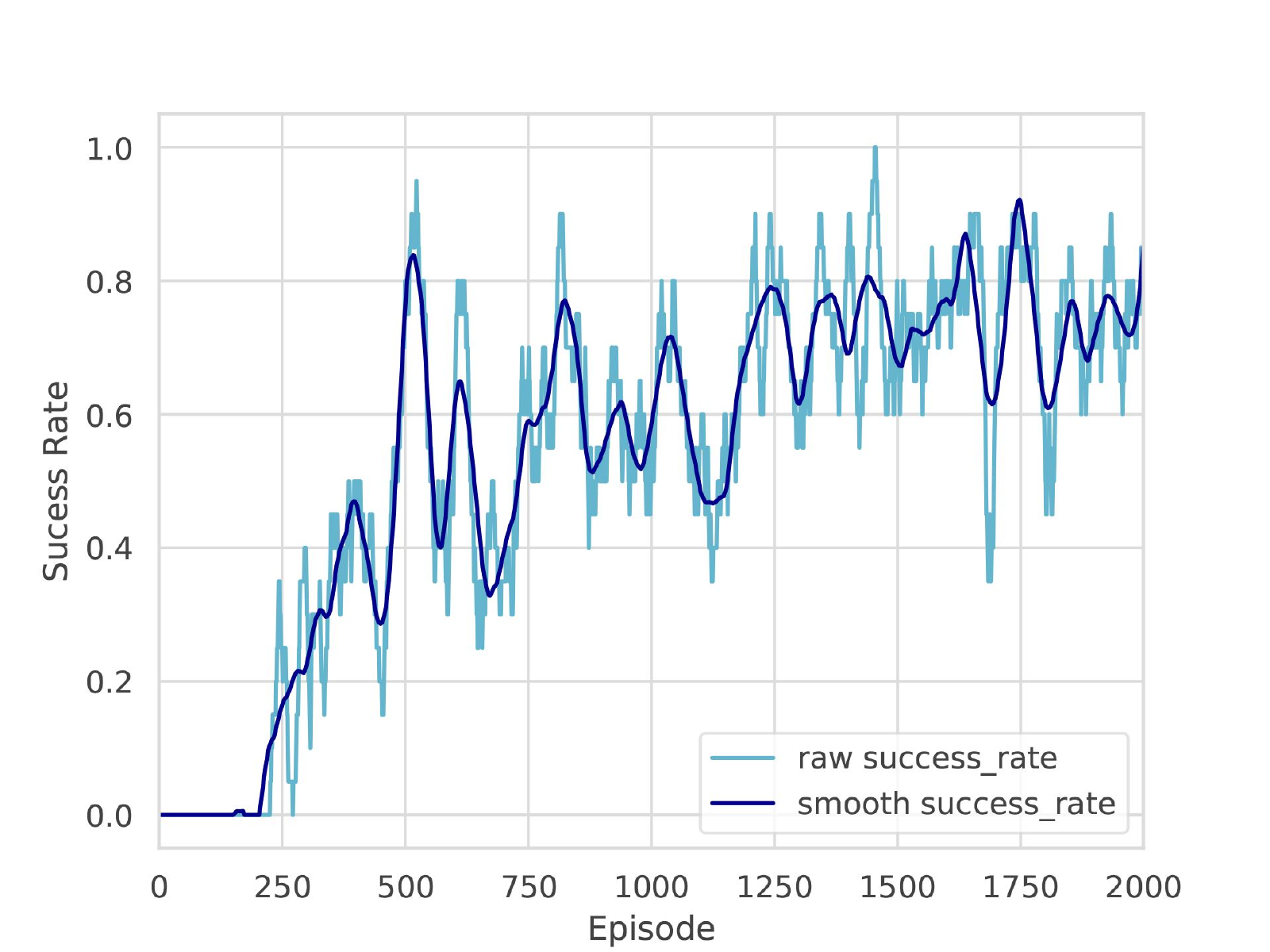}}
	  %这里是空了一行，能够实现强制将四张图分成两行两列显示，而不是放不下图了再换行，使用\\也行。
	\subfigure[Reward]{
		\label{map1_11_24reward}
		\includegraphics[width=6cm]{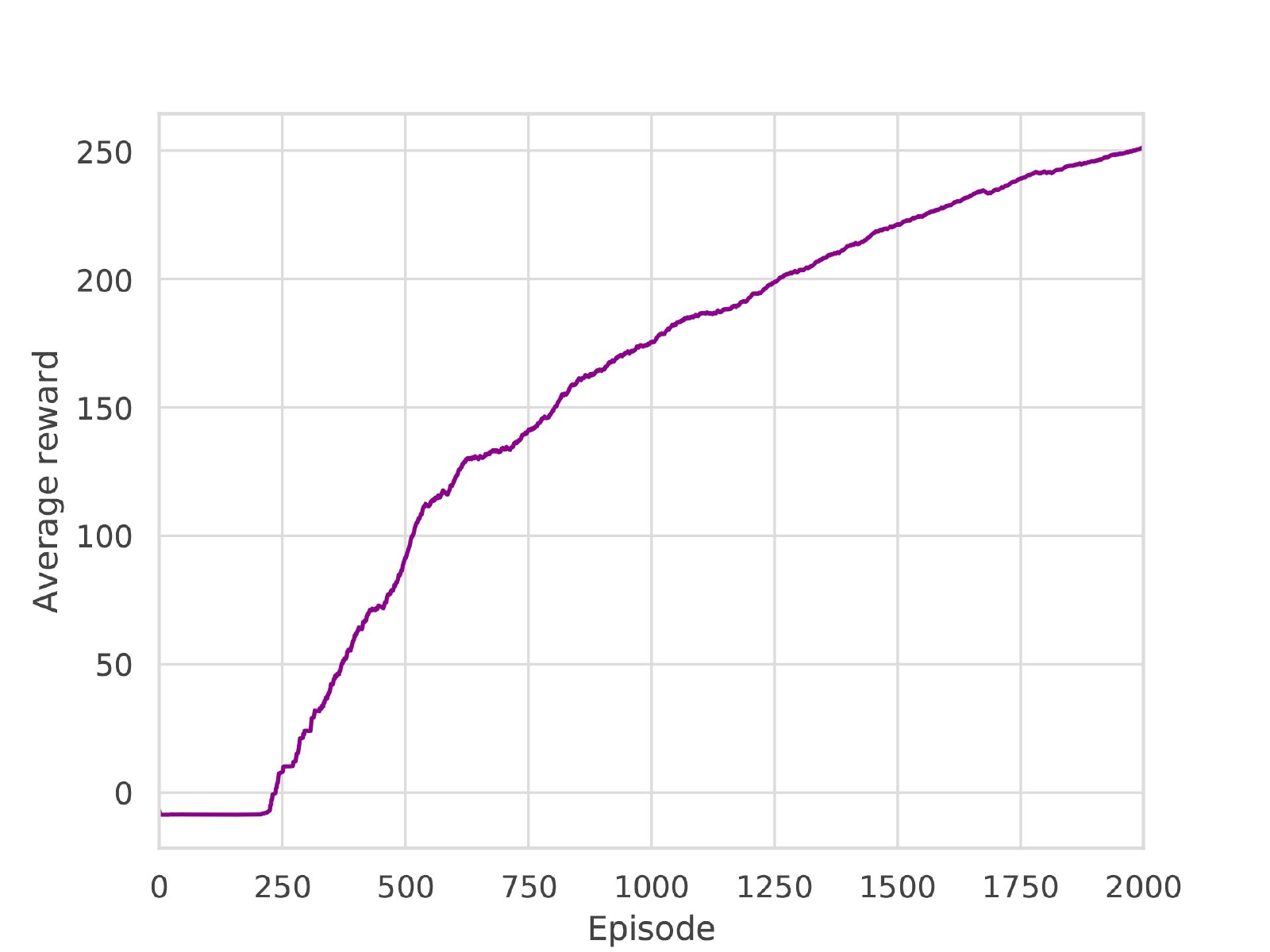}}
	\quad
    \quad
	\subfigure[Error]{
		\label{map1_11_24error}
		\includegraphics[width=6cm]{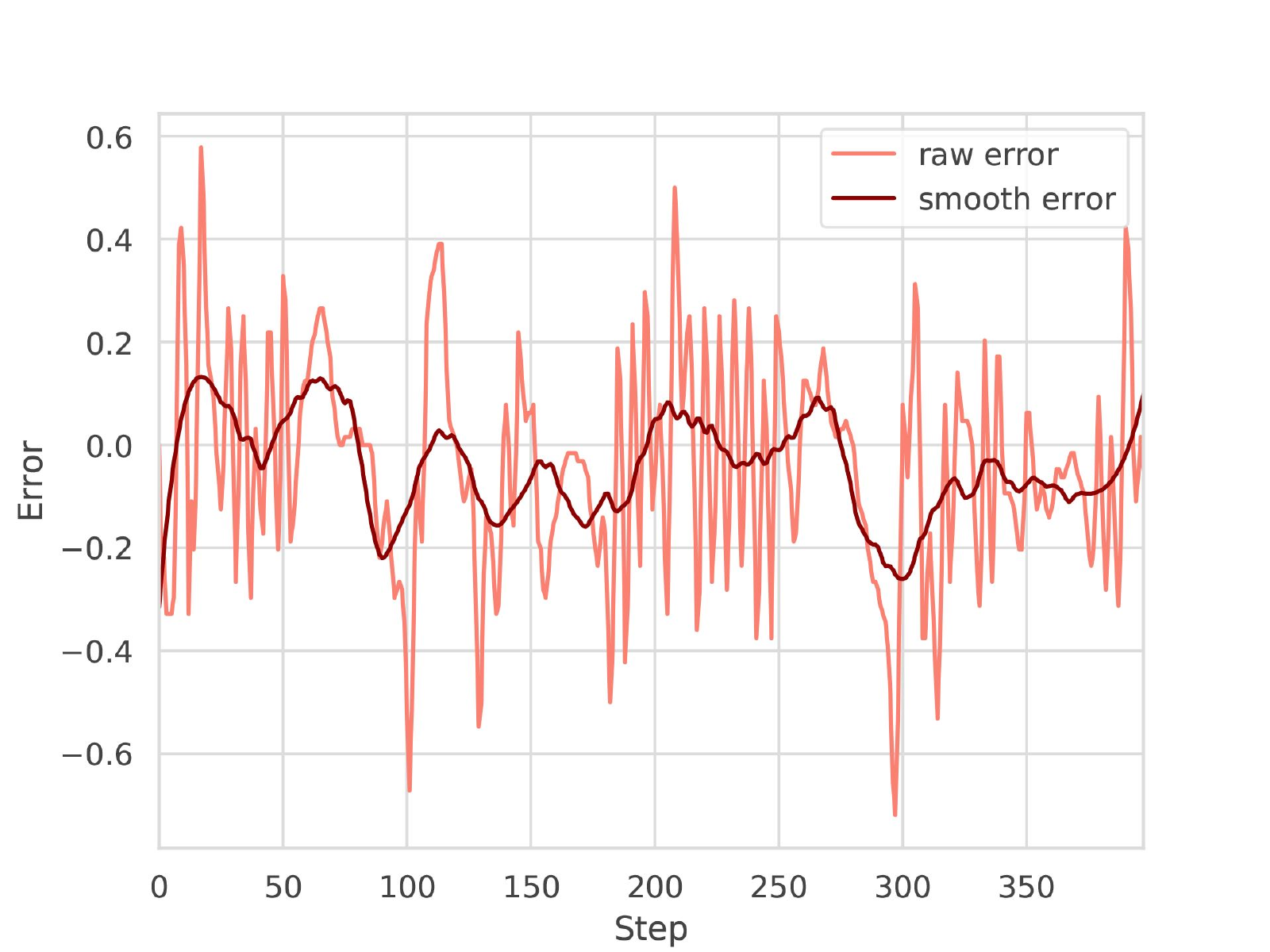}}
	\caption{Path 1 and corresponding performance parameters curves}
	\label{The performance of map1}
\end{figure}

\begin{figure}[H] %这里使用的是强制位置，除非真的放不下，不然就是写在哪里图就放在哪里，不会乱动
	\centering  %图片全局居中
	\vspace{-0.35cm} %设置与上面正文的距离
	\subfigtopskip=2pt %设置子图与上面正文或别的内容的距离
	\subfigbottomskip=3pt %设置第二行子图与第一行子图的距离，即下面的头与上面的脚的距离
	\subfigcapskip=-5pt %设置子图与子标题之间的距离
	\subfigure[Path 2 and trajectory]{
		\label{map4_track}
		\includegraphics[scale=0.09]{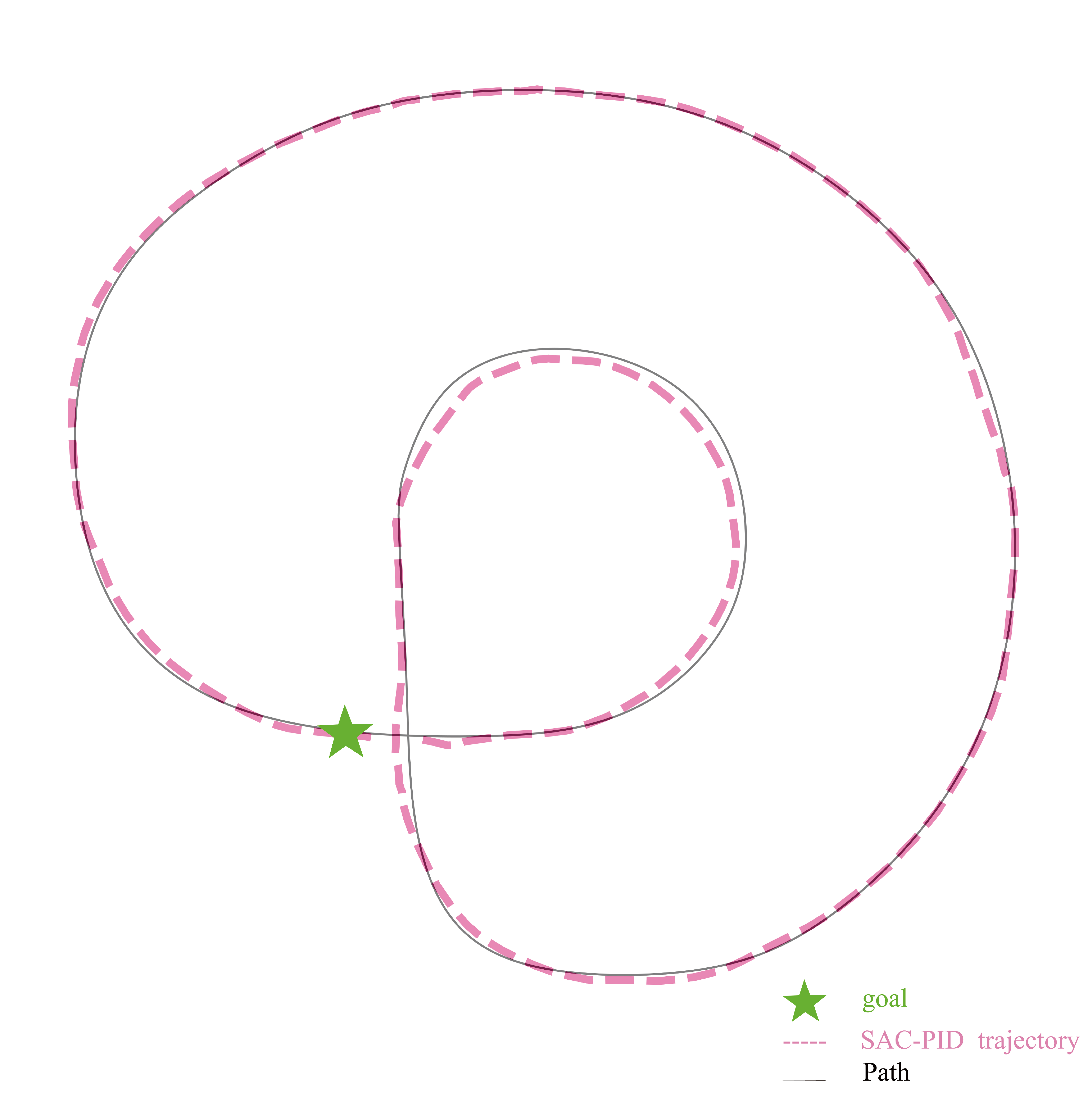}}
	\qquad %默认情况下两个子图之间空的较少，使用这个命令加大宽度
	\qquad %默认情况下两个子图之间空的较少，使用这个命令加大宽度
    \quad
	\subfigure[Success rate]{
		\label{map4_11_21success_rate}
		\includegraphics[width=6cm]{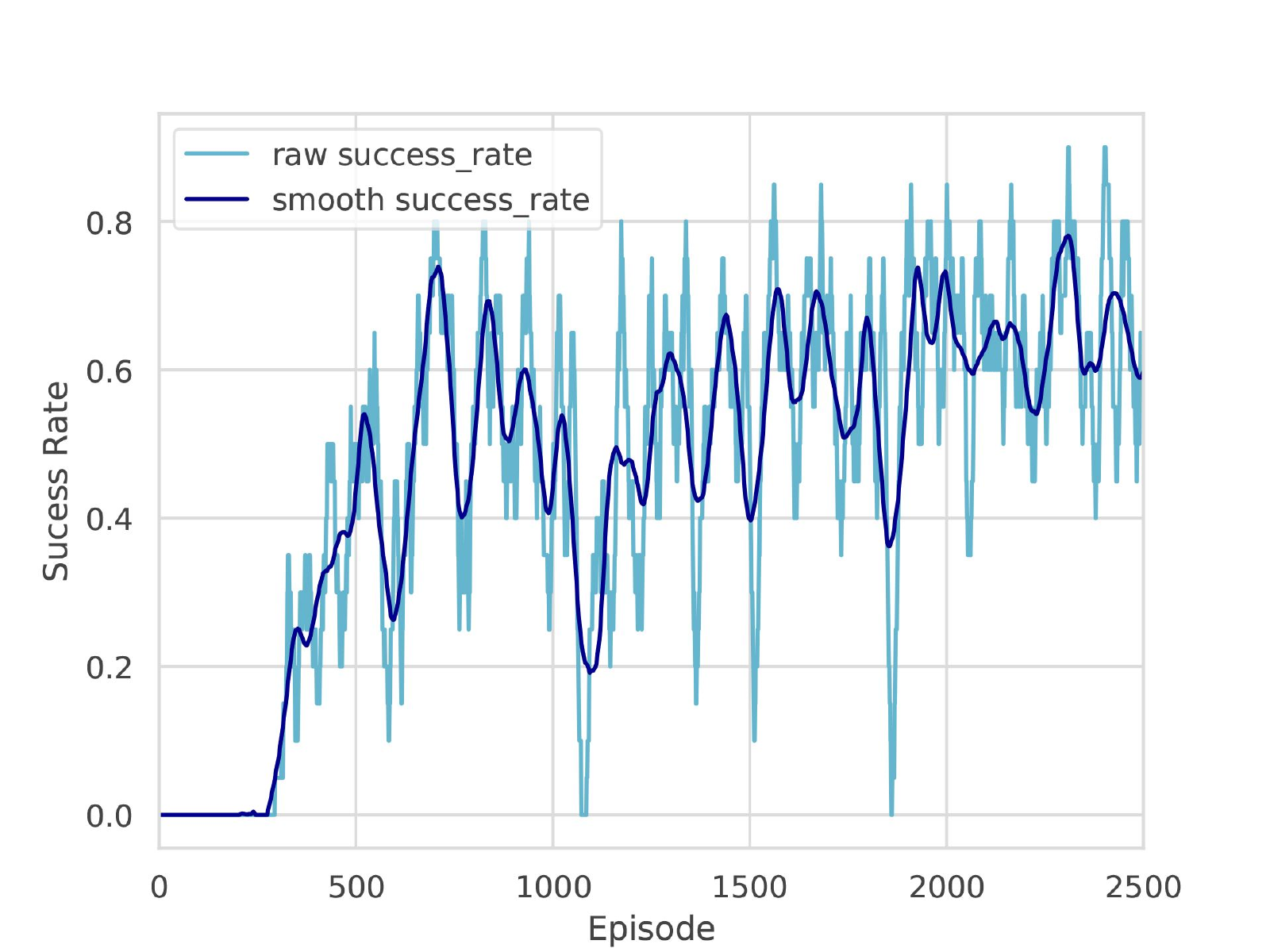}}

	\subfigure[Reward]{
		\label{map4_11_21reward}
		\includegraphics[width=6cm]{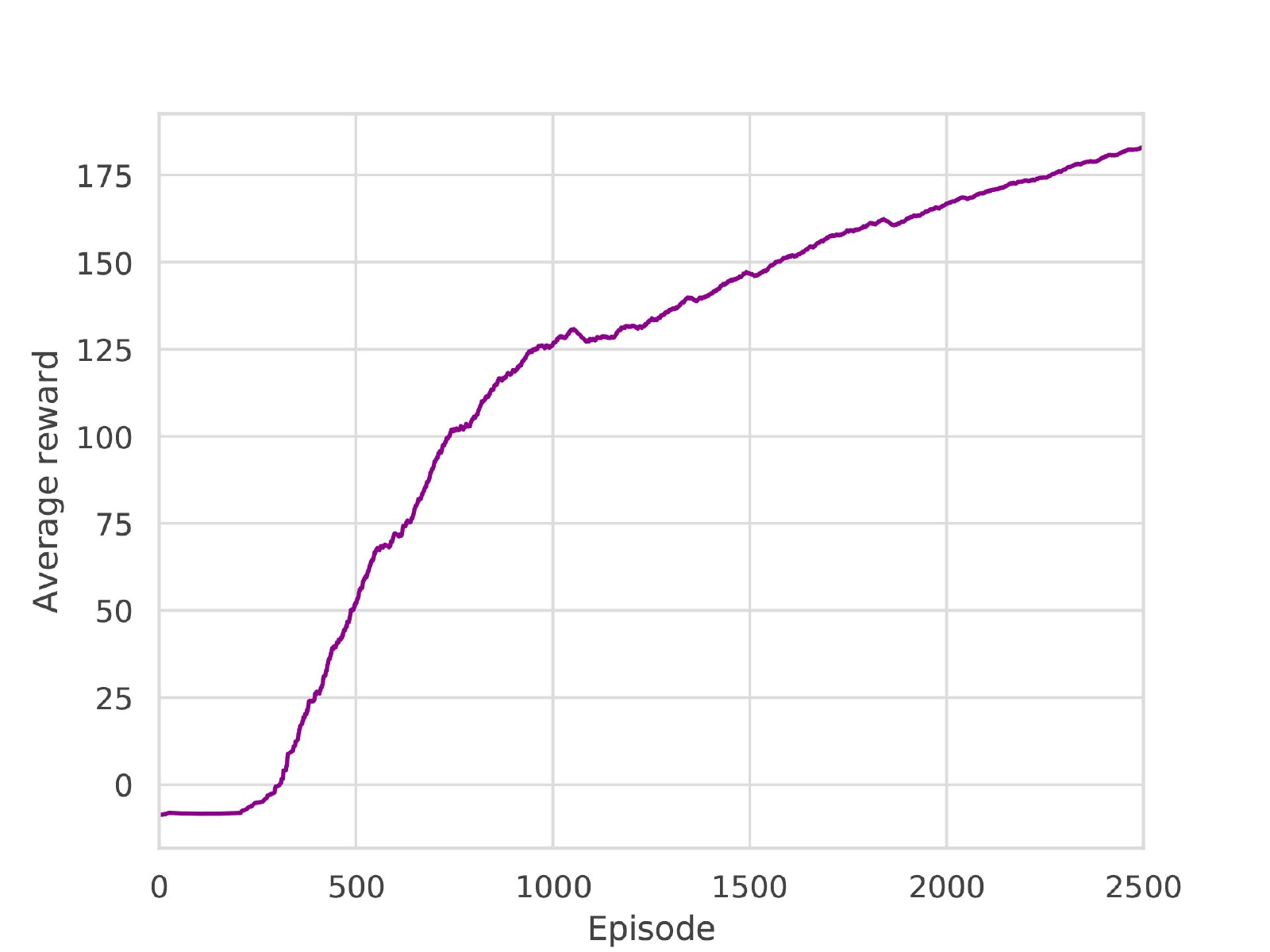}}
	\subfigure[Error]{
		\label{map4_11_21error}
		\includegraphics[width=6cm]{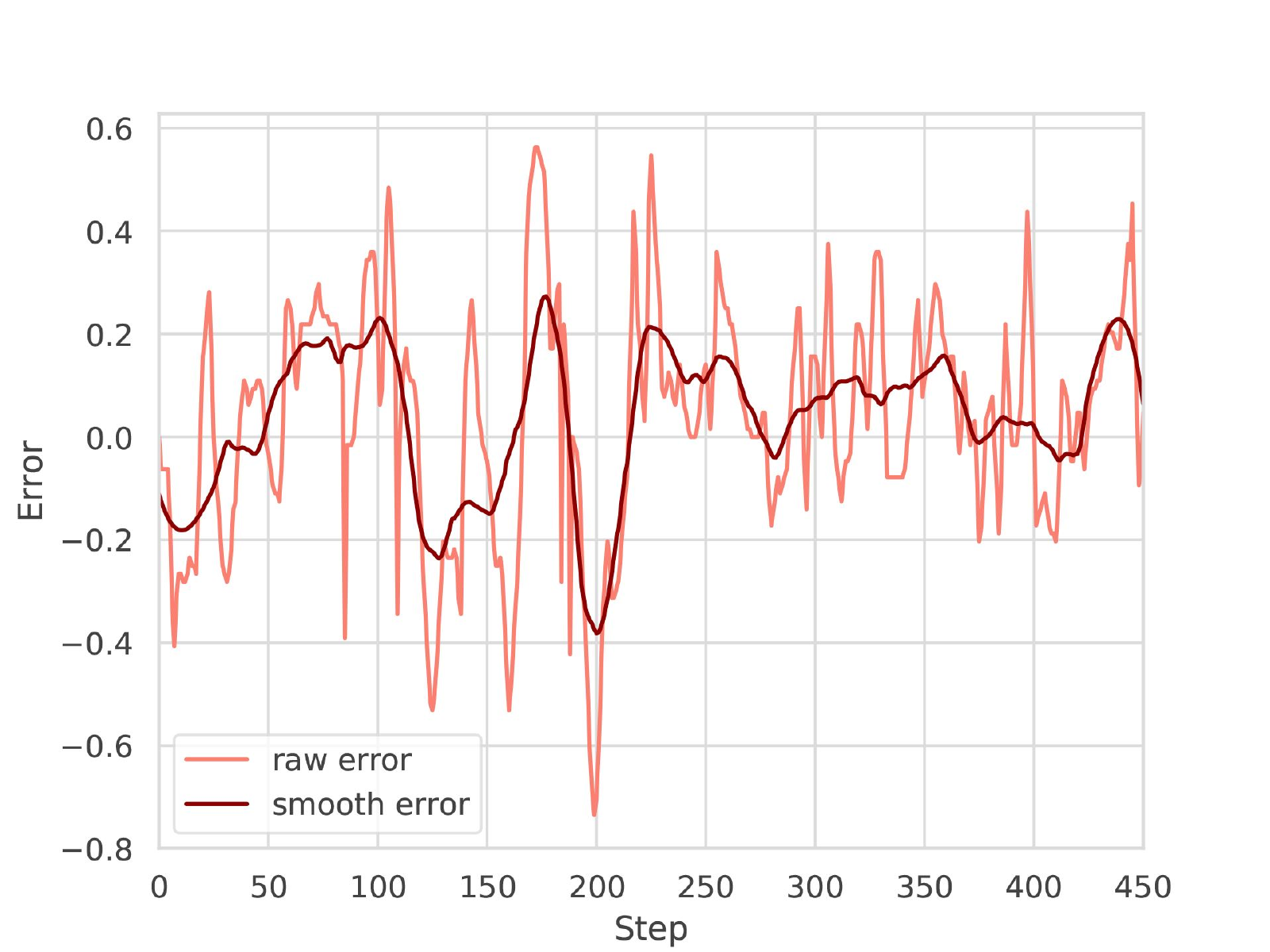}}
	\caption{Path 2 and corresponding performance parameters curves}
	\label{The performance of Map2}
\end{figure}

\begin{figure}[H] %这里使用的是强制位置，除非真的放不下，不然就是写在哪里图就放在哪里，不会乱动
	\centering  %图片全局居中
	\vspace{-0.35cm} %设置与上面正文的距离
	\subfigtopskip=2pt %设置子图与上面正文或别的内容的距离
	\subfigbottomskip=3pt %设置第二行子图与第一行子图的距离，即下面的头与上面的脚的距离
	\subfigcapskip=-5pt %设置子图与子标题之间的距离
	\subfigure[Path 3 and trajectory]{
		\label{map3_track}
		\includegraphics[scale=0.09]{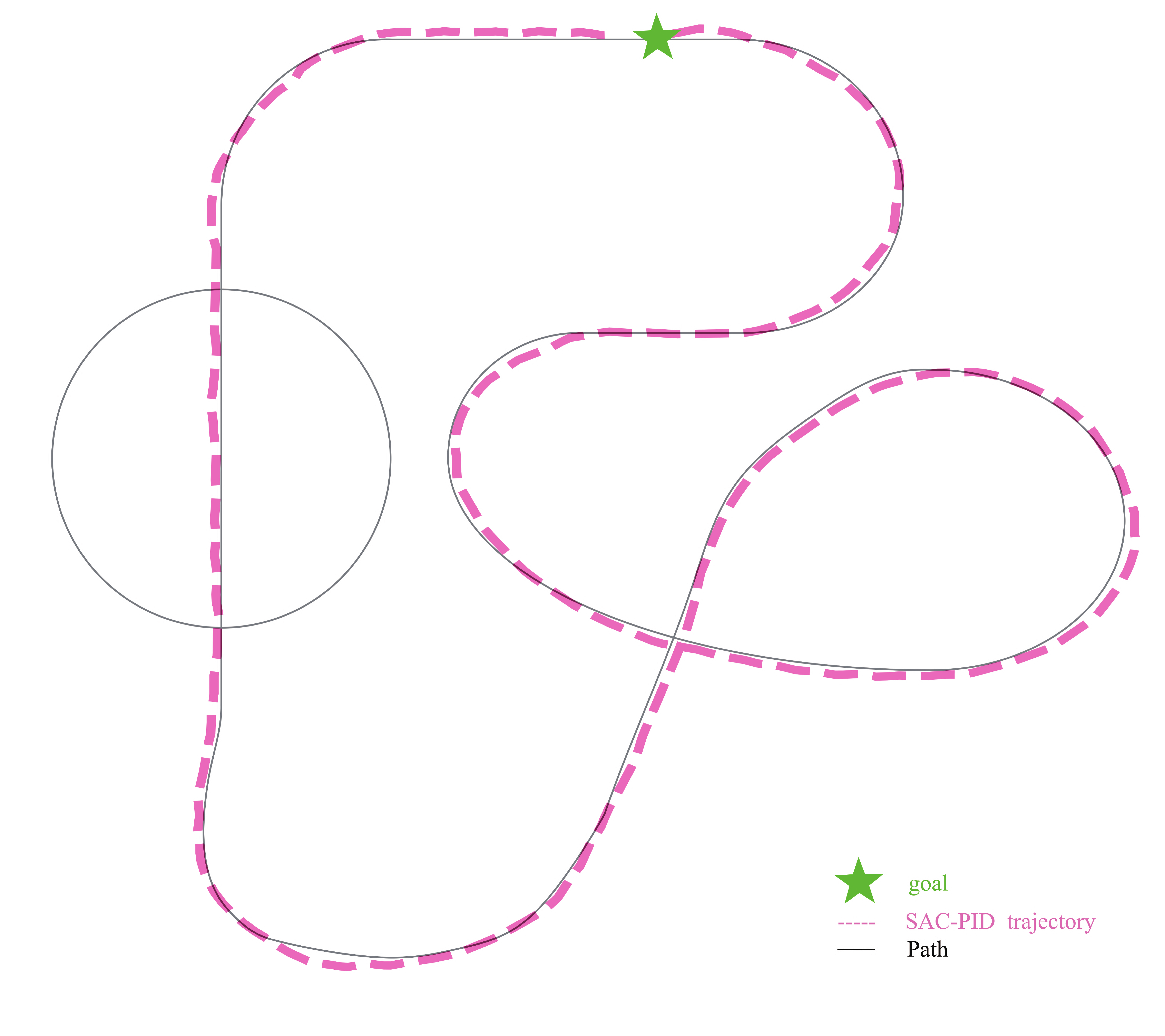}}
	\quad %默认情况下两个子图之间空的较少，使用这个命令加大宽度
	\qquad %默认情况下两个子图之间空的较少，使用这个命令加大宽度
    \quad
    \qquad
	\subfigure[Success rate]{
		\label{eval_11_21success_rate}
		\includegraphics[width=6cm]{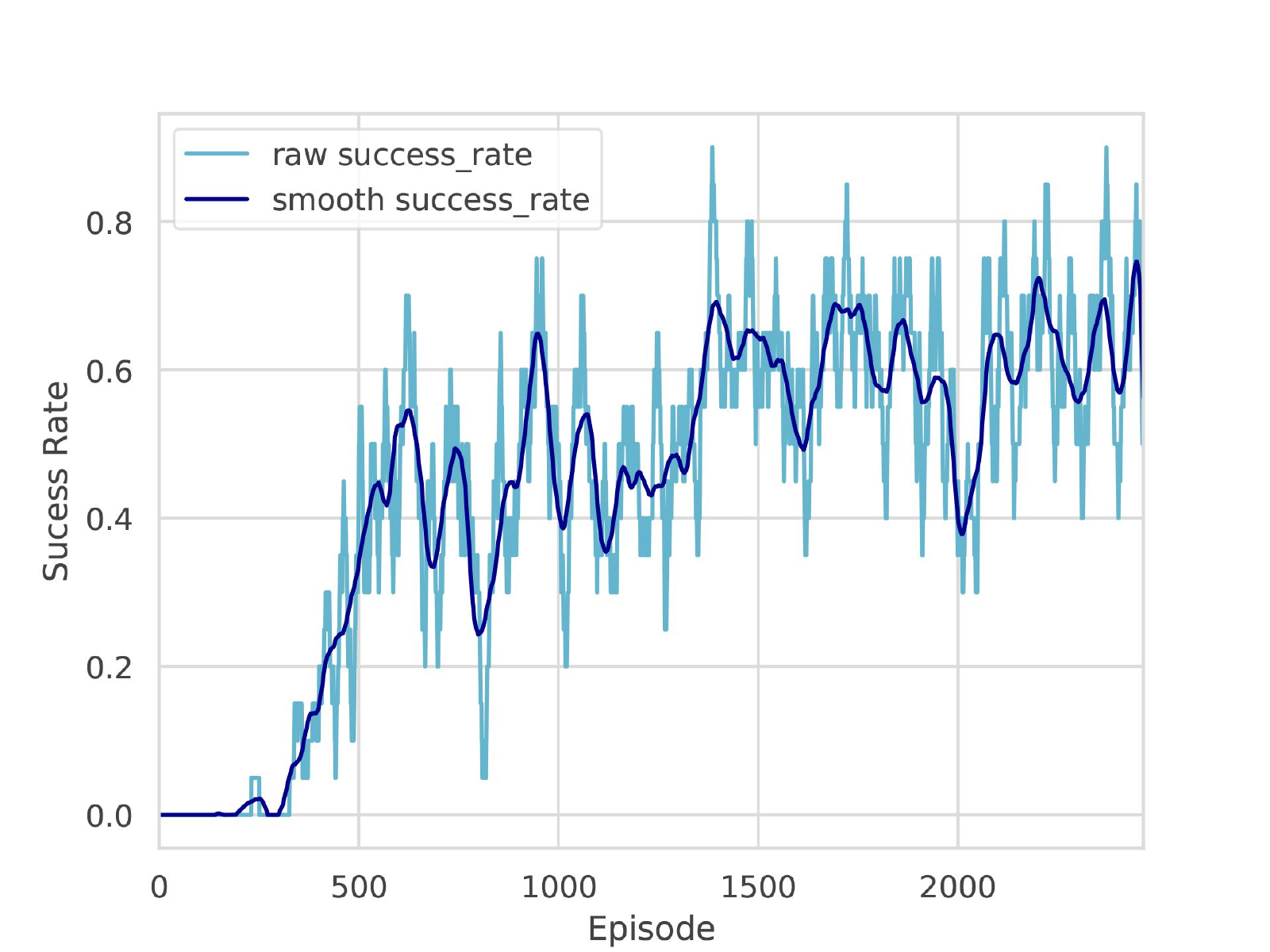}}
	  %这里是空了一行，能够实现强制将四张图分成两行两列显示，而不是放不下图了再换行，使用\\也行。
	\subfigure[Reward]{
		\label{eval_11_21reward}
		\includegraphics[width=6cm]{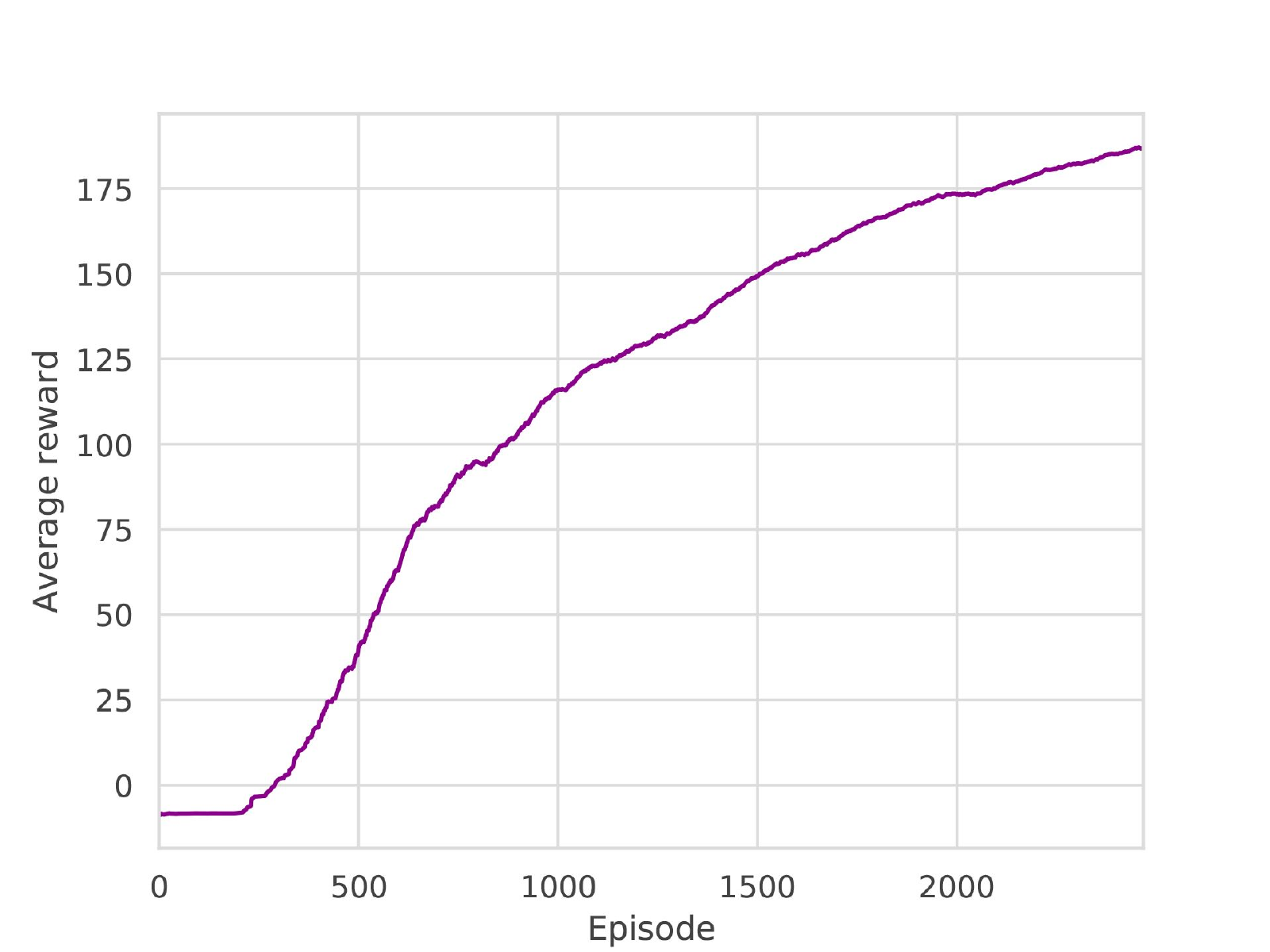}}
	\quad
    \quad
	\subfigure[Error]{
		\label{eval_11_21error}
		\includegraphics[width=6cm]{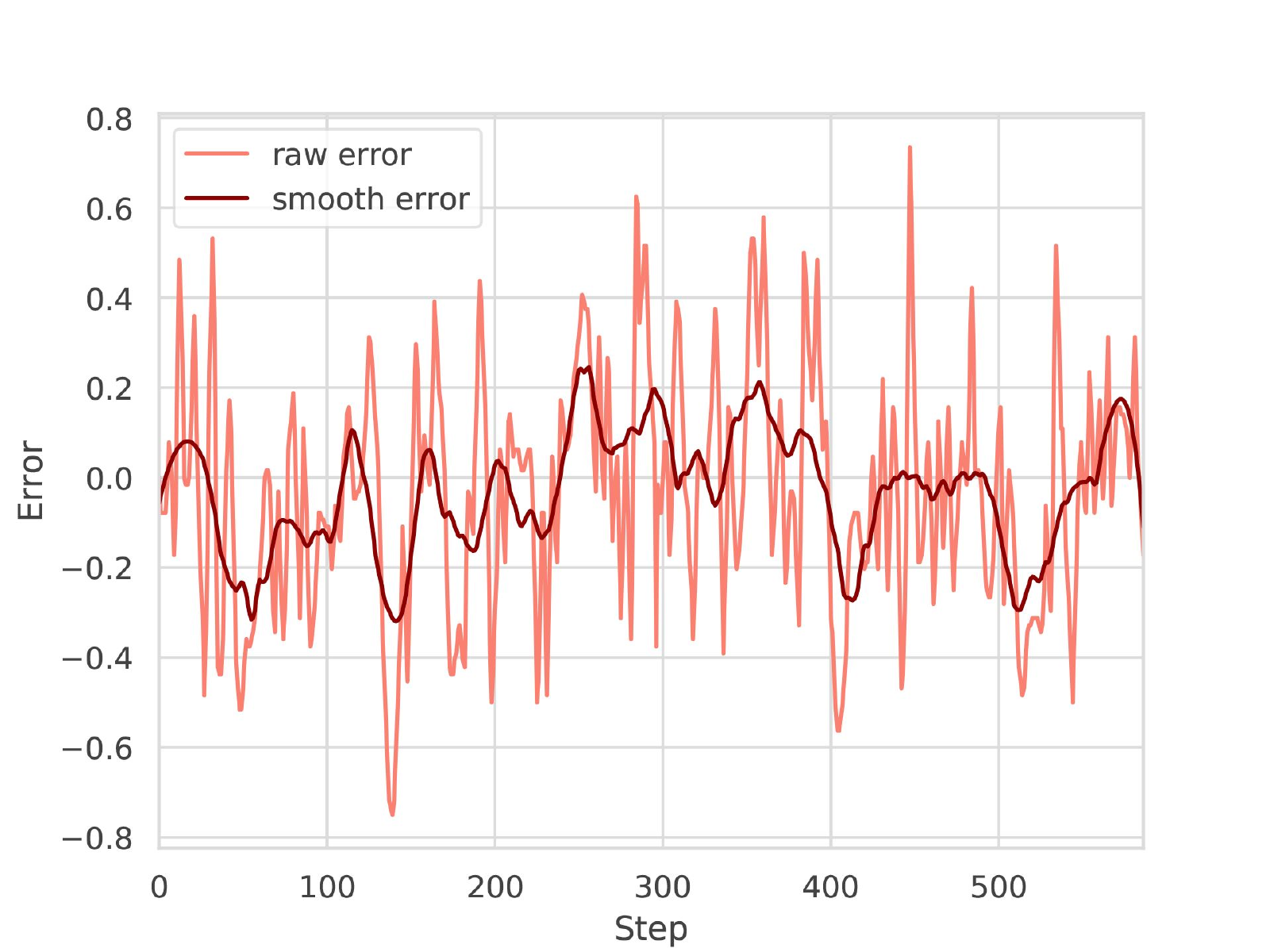}}
	\caption{Path 3 and corresponding performance parameters curves}
	\label{The performance of map3}
\end{figure}

\begin{figure}[H] %这里使用的是强制位置，除非真的放不下，不然就是写在哪里图就放在哪里，不会乱动
	\centering  %图片全局居中
	\vspace{-0.35cm} %设置与上面正文的距离
	\subfigtopskip=2pt %设置子图与上面正文或别的内容的距离
	\subfigbottomskip=3pt %设置第二行子图与第一行子图的距离，即下面的头与上面的脚的距离
	\subfigcapskip=-5pt %设置子图与子标题之间的距离
	\subfigure[Path 4 and trajectories]{
		\label{fuzzy_sac_track}
		\includegraphics[scale=0.08]{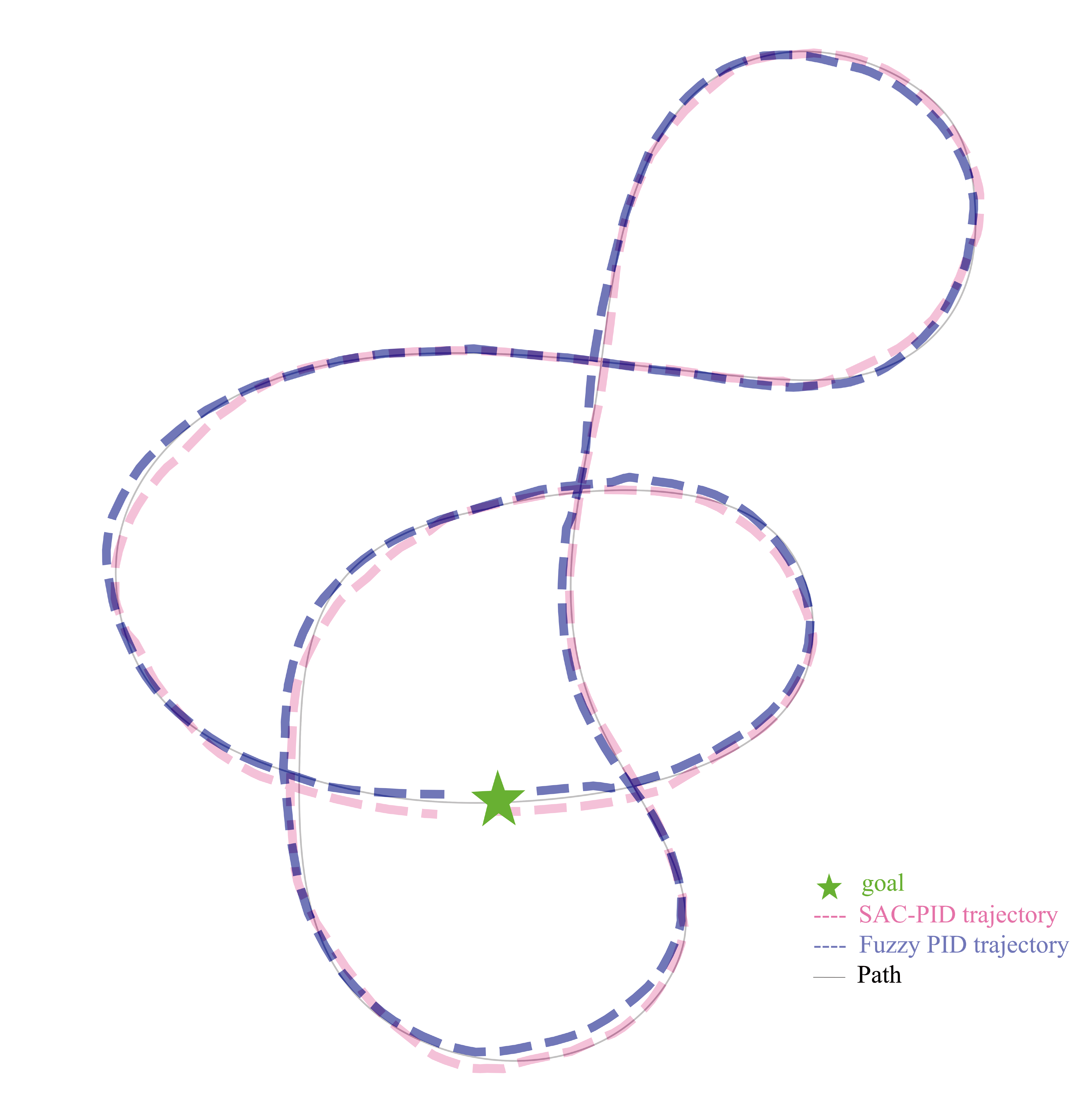}}
	\qquad %默认情况下两个子图之间空的较少，使用这个命令加大宽度
    \qquad
    \qquad
    \quad
	\subfigure[Success rate]{
		\label{map3_11_10success_rate}
		\includegraphics[width=6cm]{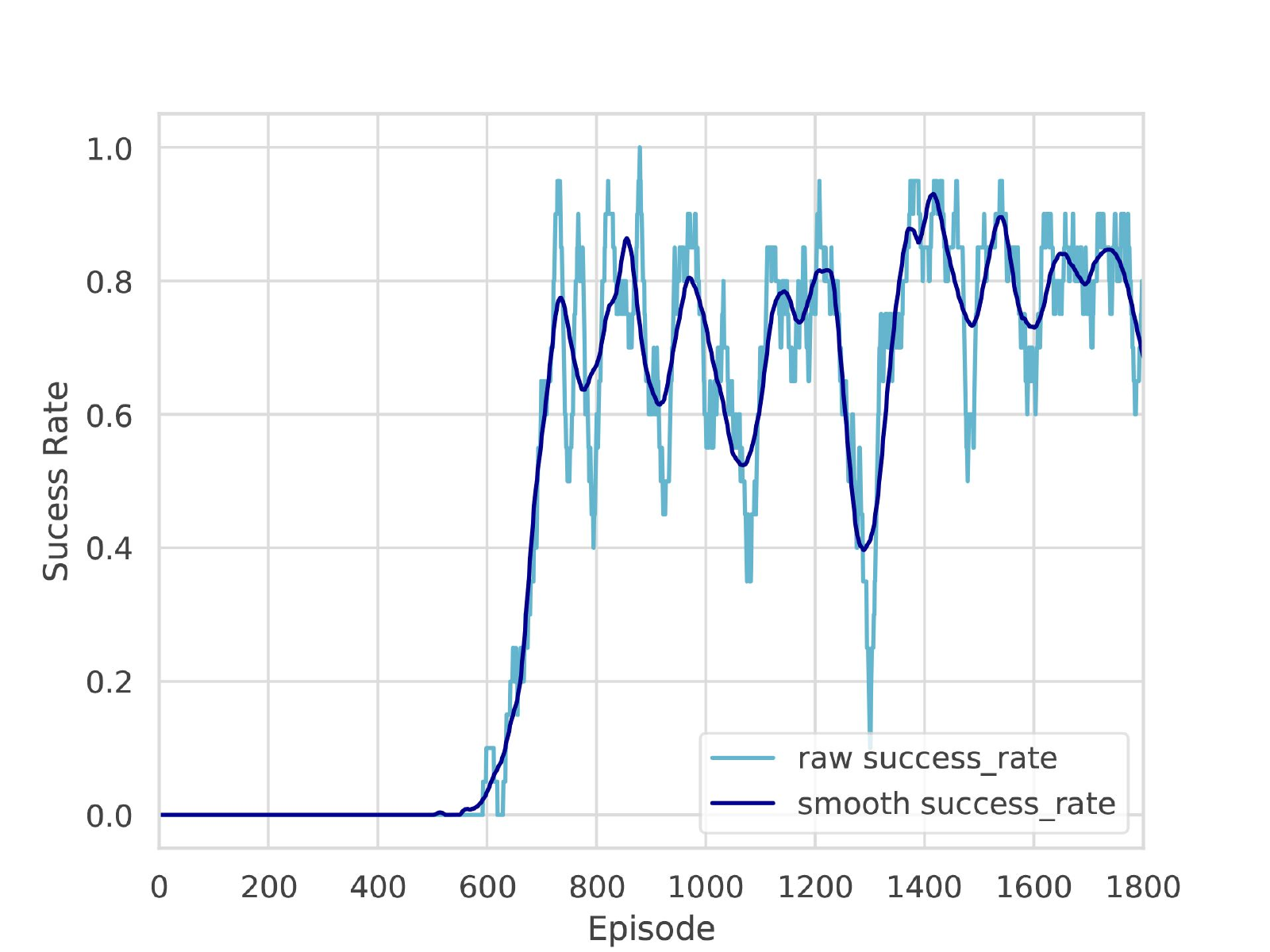}}
	  %这里是空了一行，能够实现强制将四张图分成两行两列显示，而不是放不下图了再换行，使用\\也行。
	\subfigure[Reward]{
		\label{map3_11_10reward}
		\includegraphics[width=6cm]{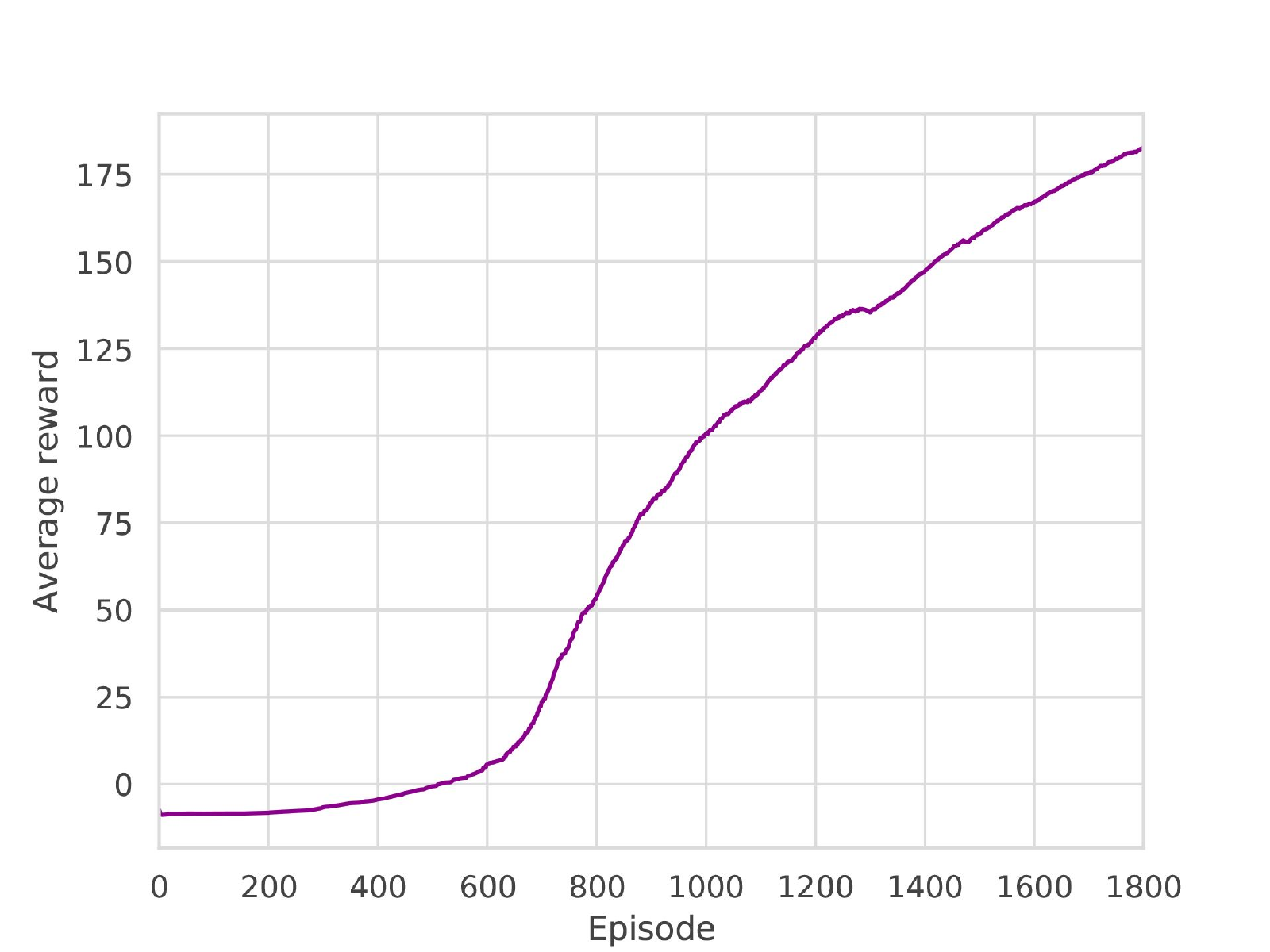}}
	\quad
	\subfigure[Error]{
		\label{map3_11_10error}
		\includegraphics[width=6cm]{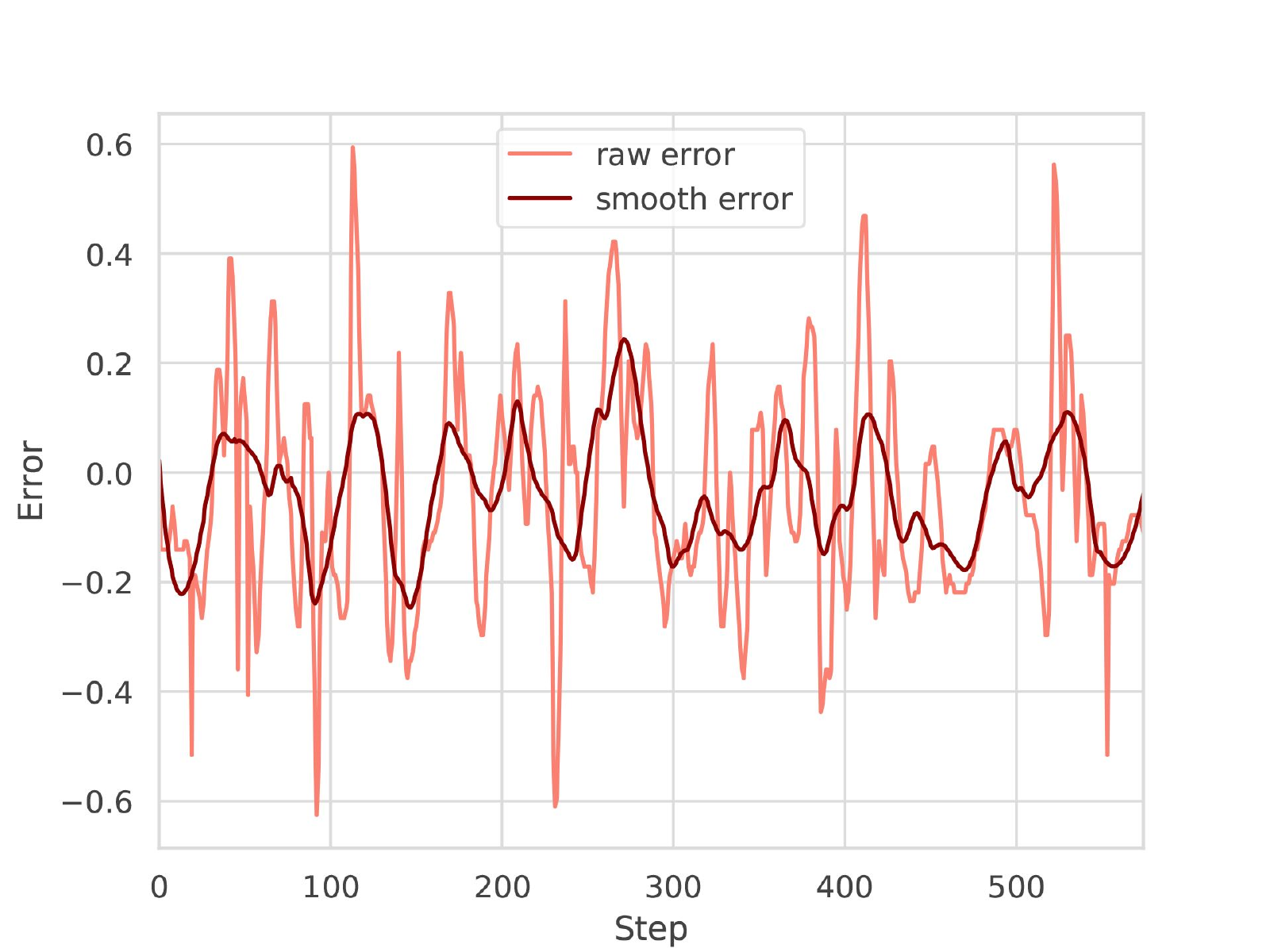}}
	\caption{Path 4 and corresponding performance parameters curves}
	\label{The performance of Map4}
\end{figure}

The whole process of the experiment is divided into two stages: the training phase and the testing phase.
We randomly select four different difficulty paths, namely playground circle (Path 1) shown in Fig. \ref{map1_track}, circle with one fork (Path 2) shown in Fig. \ref{map4_track}, complex circle with some forks (Path 3) shown in Fig. \ref{map3_track} and complex circle with many forks (Path 4) shown in Fig. \ref{fuzzy_sac_track}. The start and goal points of each path are the same location, which forms a loop.  In the training phase, the line-following robot was initialized to the starting point at the beginning of each episode. If the mobile robot reaches the goal or exits the line, it enters the next episode. The success rate is employed  as an evaluation index, the number of success times that the robot reached its goal without out of line over the total testing numbers. If the success rates of multiple episodes can stabilize to the same level, the training process is completed. After the training of each path is finished, the error data $x_4$ is served as the evaluation index of the SAC-PID control system stability. In addition, average reward of each episode can also evaluate the stability of the algorithm. In order to make it easier for readers to observe the experimental results, we smoothed raw data of error and success rate respectively to generate smooth error and smooth success rate, which was shown in Figs. \ref{The performance of map1}-\ref{The performance of Map4}.

\begin{table}[htp]\normalsize
\setlength{\belowcaptionskip}{-0.5cm}
\setlength{\abovecaptionskip}{0.cm}
\centering
\caption{The Evaluation results of each path in Gazebo platform}
\begin{tabular}{cccccc}
\toprule
\makecell[c]{Path}   &\makecell[c]{Model} & \makecell[c]{Testing number} & \makecell[c]{Success number} & \makecell[c]{Success rate(\%)}  &\makecell[c]{Average velocity(m/s)}\\
\midrule
\makecell[c]{Path 1} &\makecell[c]{Model 1}  & \makecell[c]{20}              & \makecell[c]{20}                                 & \makecell[c]{100}  & \makecell[c]{0.260$ \pm$0.010}\\
                            \makecell[c]{Path 2} &\makecell[c]{Model 2}  & \makecell[c]{20}             & \makecell[c]{20}                                 & \makecell[c]{100}   & \makecell[c]{0.255$ \pm$0.004}                                \\
                            \makecell[c]{Path 3} &\makecell[c]{Model 3}  & \makecell[c]{20}             & \makecell[c]{17}                                 & \makecell[c]{85}    & \makecell[c]{0.251$ \pm$0.017}                                \\
                            \makecell[c]{Path 4} &\makecell[c]{Model 4}  & \makecell[c]{20}             & \makecell[c]{19}                                 & \makecell[c]{95}    & \makecell[c]{0.259$ \pm$0.005}                                \\
                            \makecell[c]{Path 1} &\makecell[c]{Model 3} & \makecell[c]{20}              & \makecell[c]{18}                                 & \makecell[c]{90}    & \makecell[c]{0.254$ \pm$0.009}                              \\
                            \makecell[c]{Path 2} &\makecell[c]{Model 3} & \makecell[c]{20}             & \makecell[c]{20}                                 & \makecell[c]{100}    & \makecell[c]{0.248$ \pm$0.015}                              \\
                            \makecell[c]{Path 4} &\makecell[c]{Model 3} & \makecell[c]{20}             & \makecell[c]{16}                                 & \makecell[c]{80}     & \makecell[c]{0.255$ \pm$0.009}                               \\
                            \makecell[c]{Test Path} &\makecell[c]{Model 3} & \makecell[c]{20}             & \makecell[c]{17}                                 & \makecell[c]{85}     & \makecell[c]{0.241$ \pm$0.004}\\
\bottomrule
\end{tabular}
\label{The results of each path in Gazebo platform}
\end{table}

%\begin{table}[htbp]
%\setlength{\belowcaptionskip}{-0.5cm}
%\setlength{\abovecaptionskip}{0.cm}
%\centering
%\caption{The success rate of Model 3 in other paths}
%\begin{tabular}{p{3cm}p{3cm}p{3.5cm}p{3.5cm}}
%\hline
%\makecell[c]{Path}  & \makecell[c]{Testing number} & \makecell[c]{Success number} & \makecell[c]{Success rate(\%)} \\ \hline
%\makecell[c]{Path 1} & \makecell[c]{20}              & \makecell[c]{18}                                 & \makecell[c]{90}                                  \\
%\makecell[c]{Path 2} & \makecell[c]{20}             & \makecell[c]{20}                                 & \makecell[c]{100}                                  \\
%\makecell[c]{Path 4} & \makecell[c]{20}             & \makecell[c]{16}                                 & \makecell[c]{80}                                    \\ \hline
%\end{tabular}
%\label{map3 to other}
%\end{table}

As shown in Fig. \ref{map1_11_24success_rate}, the success rate of training on the Path 1 reaches $75\%$ and converges from about 1250th episode. As shown in Fig. \ref{map3_11_10success_rate}, although the convergence starts from 1400th episode, the success rate reaches $85\%$ for the complex Path 4. The success rates of training on Paths 2 and 3 respectively reach $72\%$ and $73\%$ as shown in Figs. \ref{map4_11_21success_rate} and \ref{eval_11_21success_rate}. The results shows that the SAC-PID control method can converge to the optimal policy in line-following tasks. %Therefore, the SAC-PID control method has strong robustness and good real-time performance.

As shown in Figs. \ref{map1_11_24reward}-\ref{map3_11_10reward}, the average reward curve of each path appears a steady upward trend without obvious decline and fluctuation, which means the training process is very stable.

The error for the mobile robot is taken as an index to measure the stability of the proposed SAC-PID control method. And the error changes with the movement of the line-following robot. Specifically, the range of error is defined as $[-1,1]$. When the error of the mobile robot is not in the above range $[-1,1]$, the mobile robot is out of the line and the line-following task fails. After
\clearpage
%map1
%map2

%map3

%map4
\clearpage

\noindent{the mobile robot completes the training process, we randomly sampled one of the episodes which successfully completes the line-following task. Then the error of each step in the sampled episode is drawn, as shown in Figs. \ref{map1_11_24error}-\ref{map3_11_10reward}. Since the error range of each path is concentrated in $[-0.6,0.6]$ and the oscillation is small, the proposed SAC-PID control approach has good stability.}

\begin{figure}[htp] %这里使用的是强制位置，除非真的放不下，不然就是写在哪里图就放在哪里，不会乱动
	\centering  %图片全局居中
	\vspace{-0.35cm} %设置与上面正文的距离
	\subfigtopskip=1pt %设置子图与上面正文或别的内容的距离
	\subfigbottomskip=3pt %设置第二行子图与第一行子图的距离，即下面的头与上面的脚的距离
	\subfigcapskip=-5pt %设置子图与子标题之间的距离
	\subfigure[Test Path]{
		\label{Test Path and trajectory}
		\includegraphics[scale=0.13]{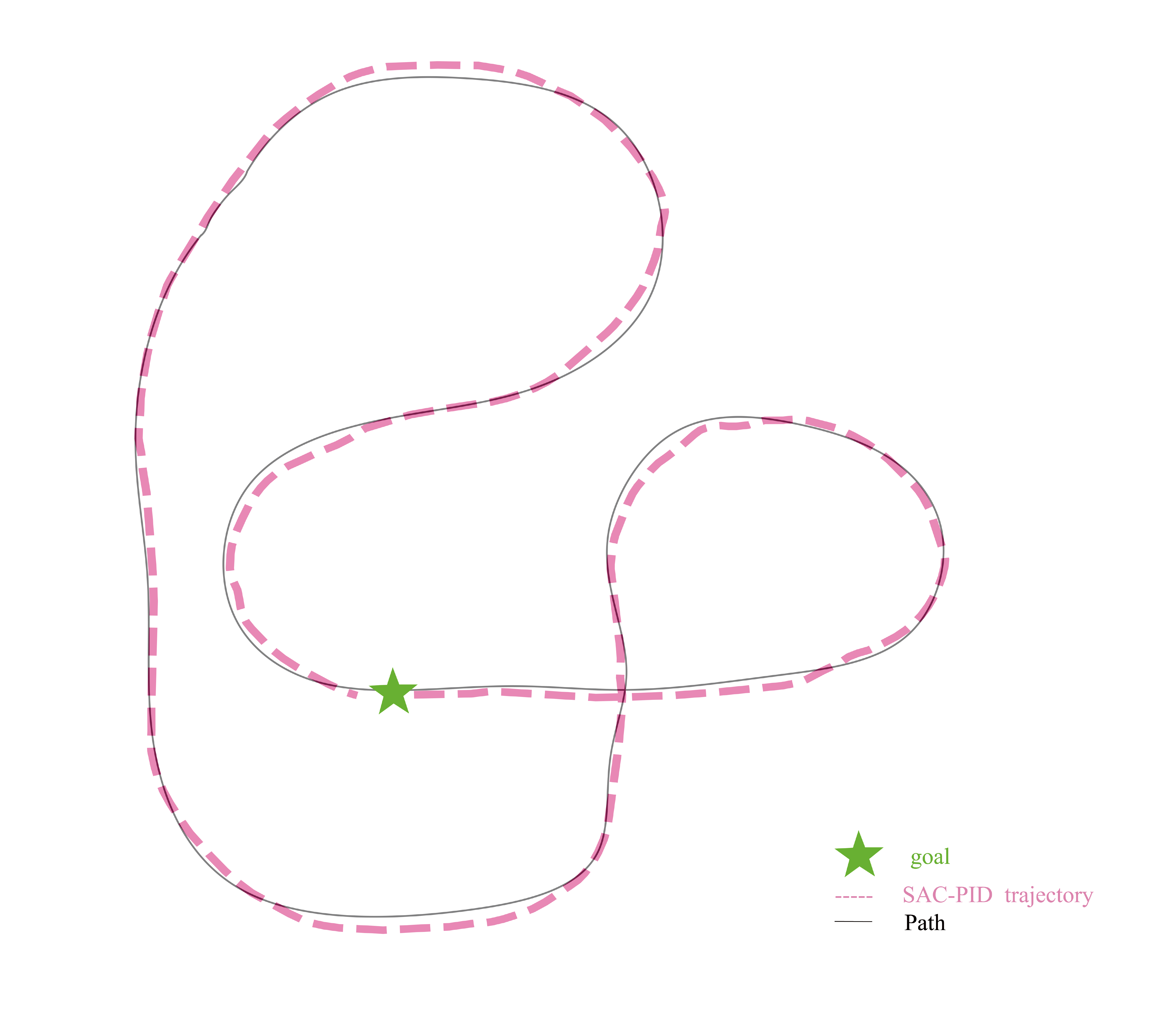}}
	\quad %默认情况下两个子图之间空的较少，使用这个命令加大宽度
 %默认情况下两个子图之间空的较少，使用这个命令加大宽度
    \quad
	\subfigure[The error for SAC-PID controller on Test Path]{
		\label{The error of SAC-PID controller on Test Path}
		\includegraphics[height=5.8cm]{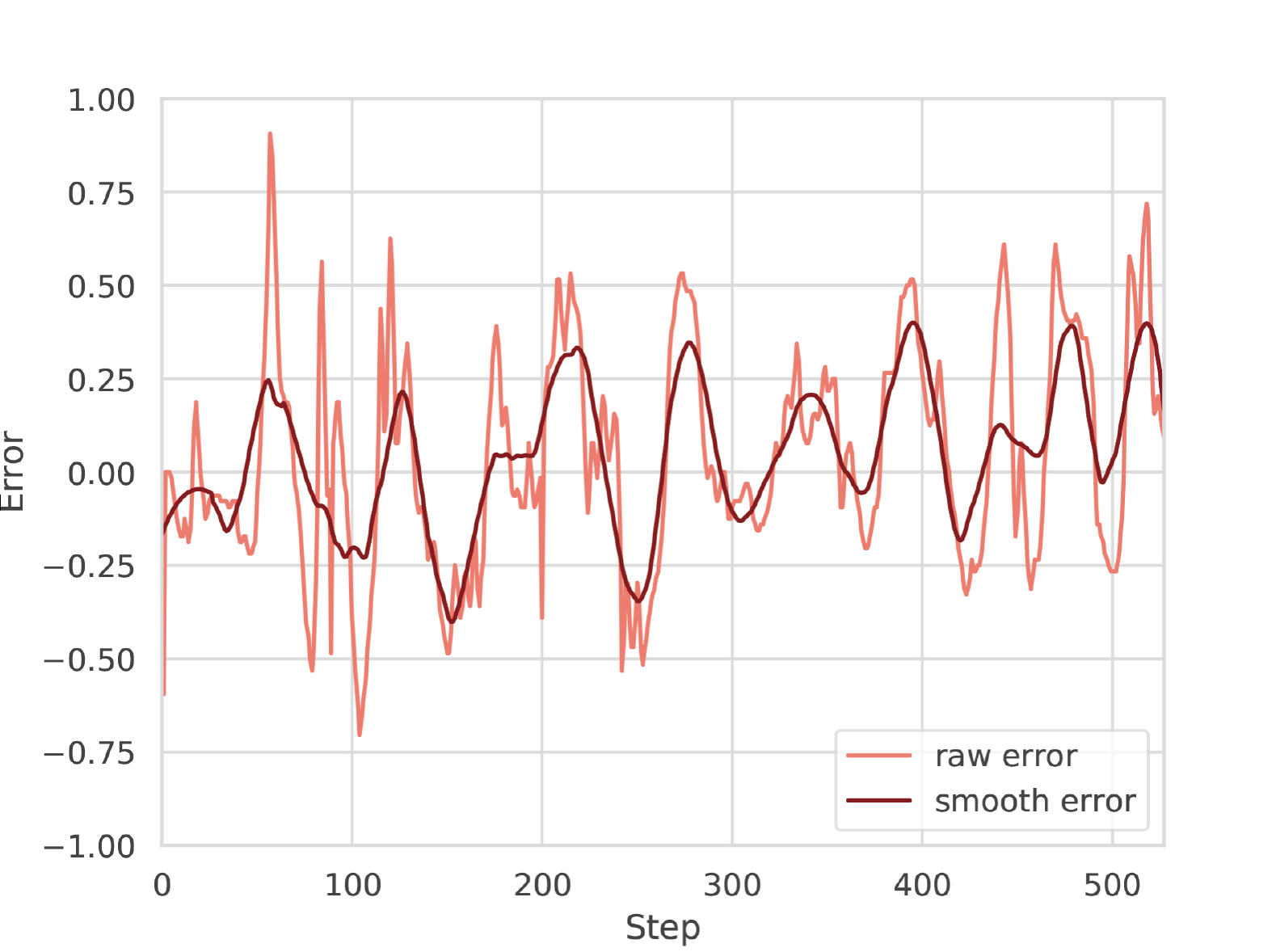}}
    \caption{Test Path and corresponding error curves}
	\label{The performance of test path}
\end{figure}

In the testing phase, we used the trained model. Specifically, the parameters of the trained model cannot be changed and the RL agent does not explore in the testing process.
Firstly, we tested four different trained models on their corresponding paths. As shown in Table \ref{The results of each path in Gazebo platform}, the success rates are $100\%$ in the simpler routes such as Paths 1 and 2. In complex routes such as Paths 3 and 4, the success rates are more than $85\%$. Then, in order to verify the generalization of the SAC-PID control method, we selected Model 3 trained in Path 3 to test all other paths with different difficulties. Furthermore, we also employed Model 3 to test the performance of SAC-PID control approach in Test Path, as shown in Fig. \ref{The performance of test path}.
As shown in Table \ref{The results of each path in Gazebo platform}, the success rates of the test results are above $80\%$, even reach $100\%$, which shows that the proposed SAC-PID control method has good generalization and strong robustness. At last, we visualised the original path and the SAC-PID tracking trajectory in Figs. \ref{map1_track}-\ref{Test Path and trajectory} to observe the effectiveness of the SAC-PID control method. It can be said with certainty that the tracking trajectory of SAC-PID fits the paths except for some paths with larger curvature.

\begin{table}[htb]
\setlength{\belowcaptionskip}{0.cm}
\setlength{\abovecaptionskip}{0.cm}
\centering
\caption{The comparison of SAC-PID control and fuzzy PID control on Path 4}
\begin{tabular}{ccccc}
\toprule
\makecell[c]{Controller}  & \makecell[c]{Testing number} & \makecell[c]{Success number} & \makecell[c]{Success rate(\%)}           & \makecell[c]{Average velocity(m/s)}\\
\midrule
\makecell[c]{SAC-PID} & \makecell[c]{20}              & \makecell[c]{19}                                 & \makecell[c]{95}        & \makecell[c]{0.259$ \pm$ 0.012}                          \\
\makecell[c]{fuzzy PID} & \makecell[c]{20}             & \makecell[c]{4}                                 & \makecell[c]{20}        & \makecell[c]{0.197 $\pm$ 0.027}                        \\
\bottomrule
\end{tabular}
\label{The comparison of SAC-PID controller and fuzzy PID controller on Path 4}
\end{table}

\subsection{Comparison with fuzzy PID control}
To the author’s best knowledge, the previous work used traditional PID controllers to complete the line-following task on the specific path. However, because of traditional PID control without the feature of adaptivity, we chose fuzzy PID control as the baseline to compare with the SAC-PID control method.  All the details of fuzzy PID control are described in Appendix \ref{appendix B}.

\begin{figure}[htp]
\centerline{\includegraphics[scale=0.57]{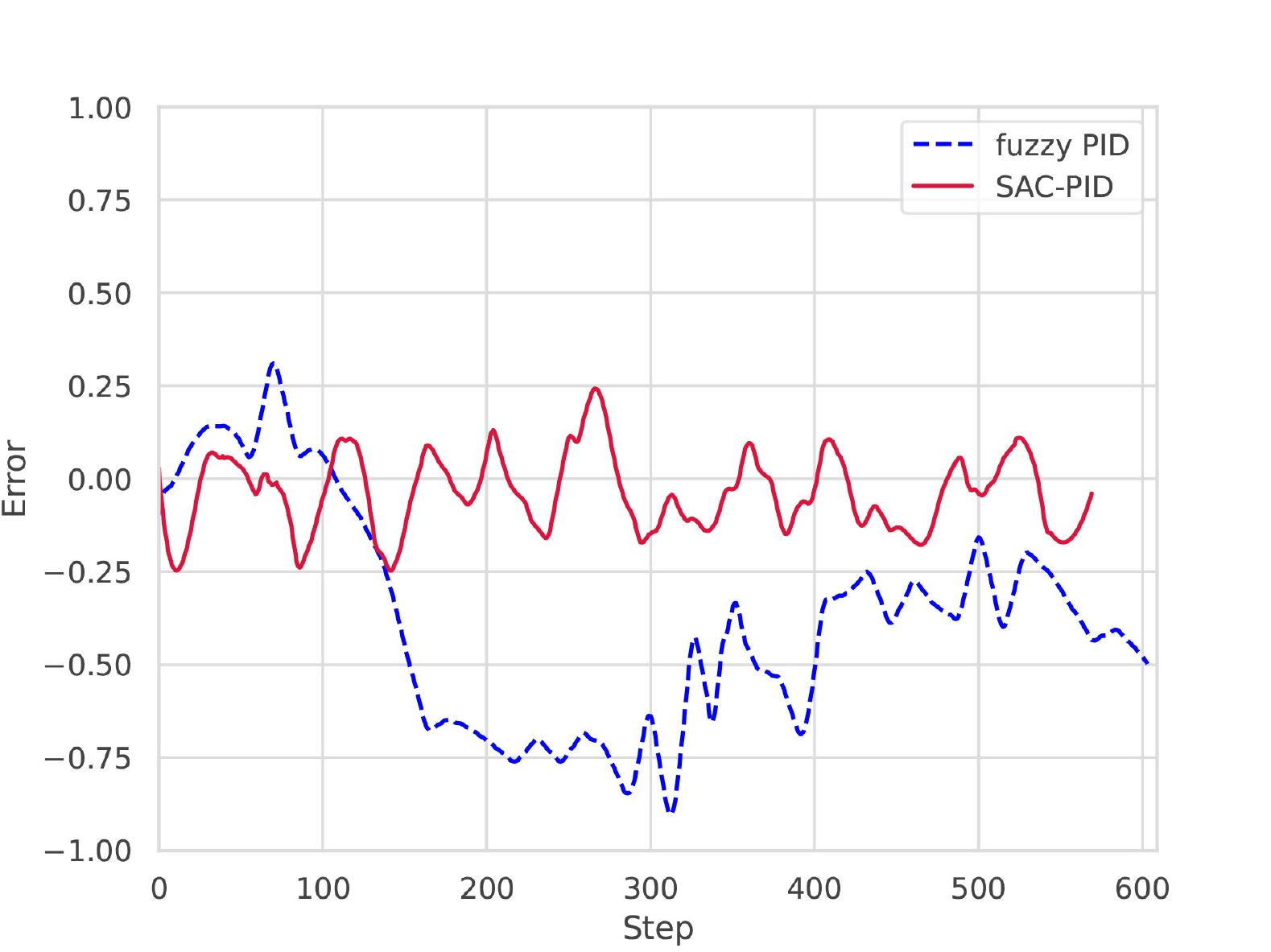}}
\caption{The error for the SAC-PID control and the fuzzy PID control\label{fuuzy_sac_com}}
\end{figure}

We chose Model 4 as the policy of SAC-PID control to compare with fuzzy PID control on the most difficult Path 4 on Gazebo. The error data is generated by the mobile robot completing the line-following task under the control of SAC-PID and fuzzy PID respectively. To facilitate comparison, the curves shown in Fig. \ref{fuuzy_sac_com} are smooth error curves. The error range of the SAC-PID control is between ${[}-0.25, 0.25{]}$, which is much smaller than that of fuzzy PID control ${[}-1, 0.25{]}$. It is obvious that the stability of SAC-PID control is significantly better than fuzzy PID control. The average linear velocity of the mobile robot controlled by the SAC-PID control is also faster than the velocity under the control of fuzzy PID control, as shown in Table \ref{The comparison of SAC-PID controller and fuzzy PID controller on Path 4}. The average velocity gap is caused by the fact that the real-time performance of SAC-PID control is much better than that of fuzzy PID control. In particular, on a computer with an i5-9400 CPU, a Python implementation of SAC-PID (Algorithm \ref{alg1}) takes about 65ms per iteration on line-following tasks. Under the same configuration, fuzzy PID control method needs 670ms per iteration. In order to compare the effectiveness of SAC-PID control and fuzzy PID control, the trajectories of the SAC-PID control and fuzzy PID control are also displayed in Fig. \ref{fuzzy_sac_track}. It is seen that the performance of the SAC-PID control is much better than that of fuzzy PID control.

\subsection{Real world tests}
\begin{figure}[htp] %这里使用的是强制位置，除非真的放不下，不然就是写在哪里图就放在哪里，不会乱动
	\centering  %图片全局居中
	\vspace{-0.35cm} %设置与上面正文的距离
	\subfigtopskip=1pt %设置子图与上面正文或别的内容的距离
	\subfigbottomskip=3pt %设置第二行子图与第一行子图的距离，即下面的头与上面的脚的距离
	\subfigcapskip=-5pt %设置子图与子标题之间的距离
	\subfigure[Real path 1]{
		\label{map1_real}
		\includegraphics[scale=0.052]{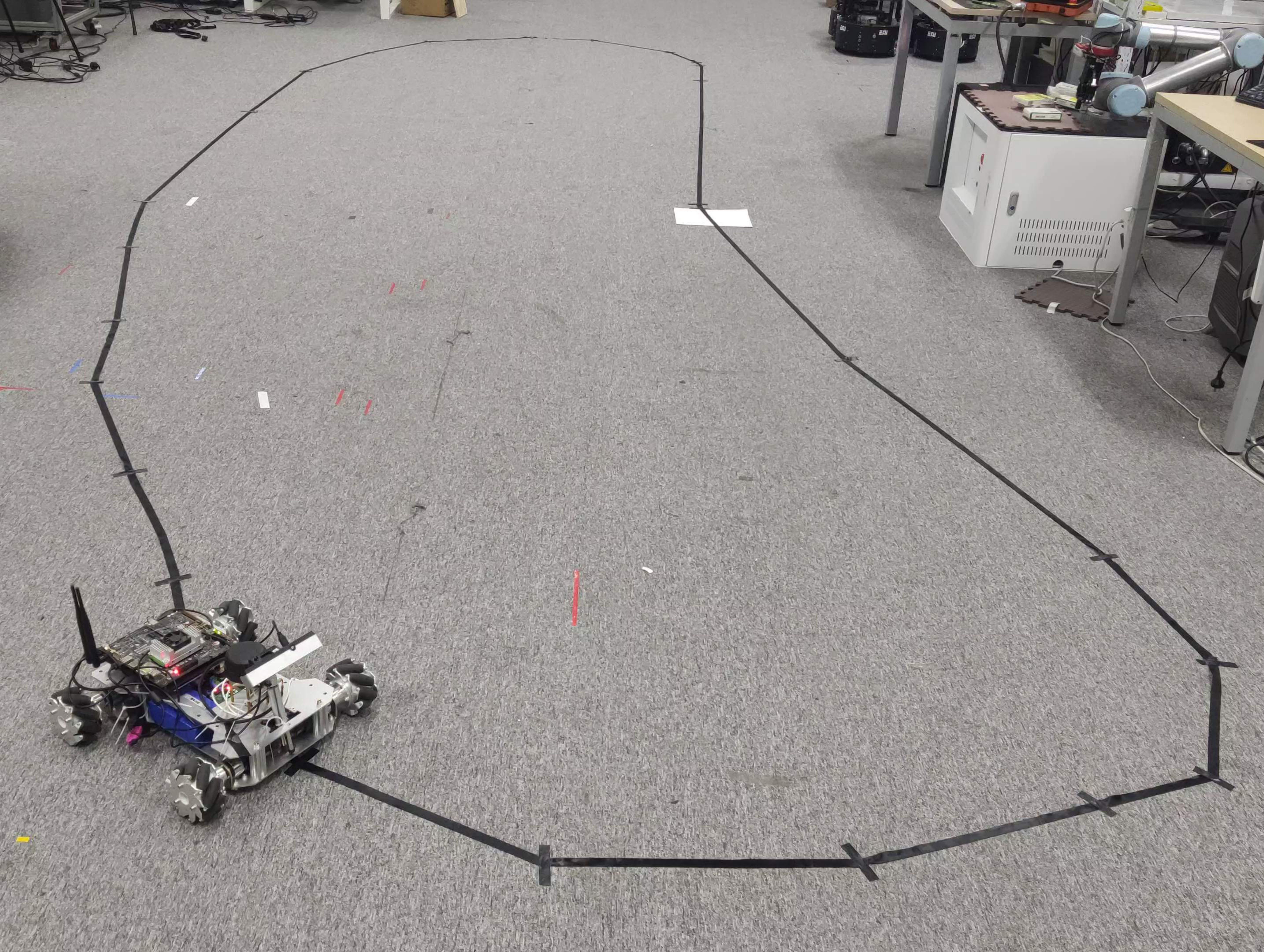}}
	\quad %默认情况下两个子图之间空的较少，使用这个命令加大宽度
 %默认情况下两个子图之间空的较少，使用这个命令加大宽度
    \quad
	\subfigure[The error of SAC-PID controller on Real path 1]{
		\label{map1_error_real}
		\includegraphics[height=4.5cm]{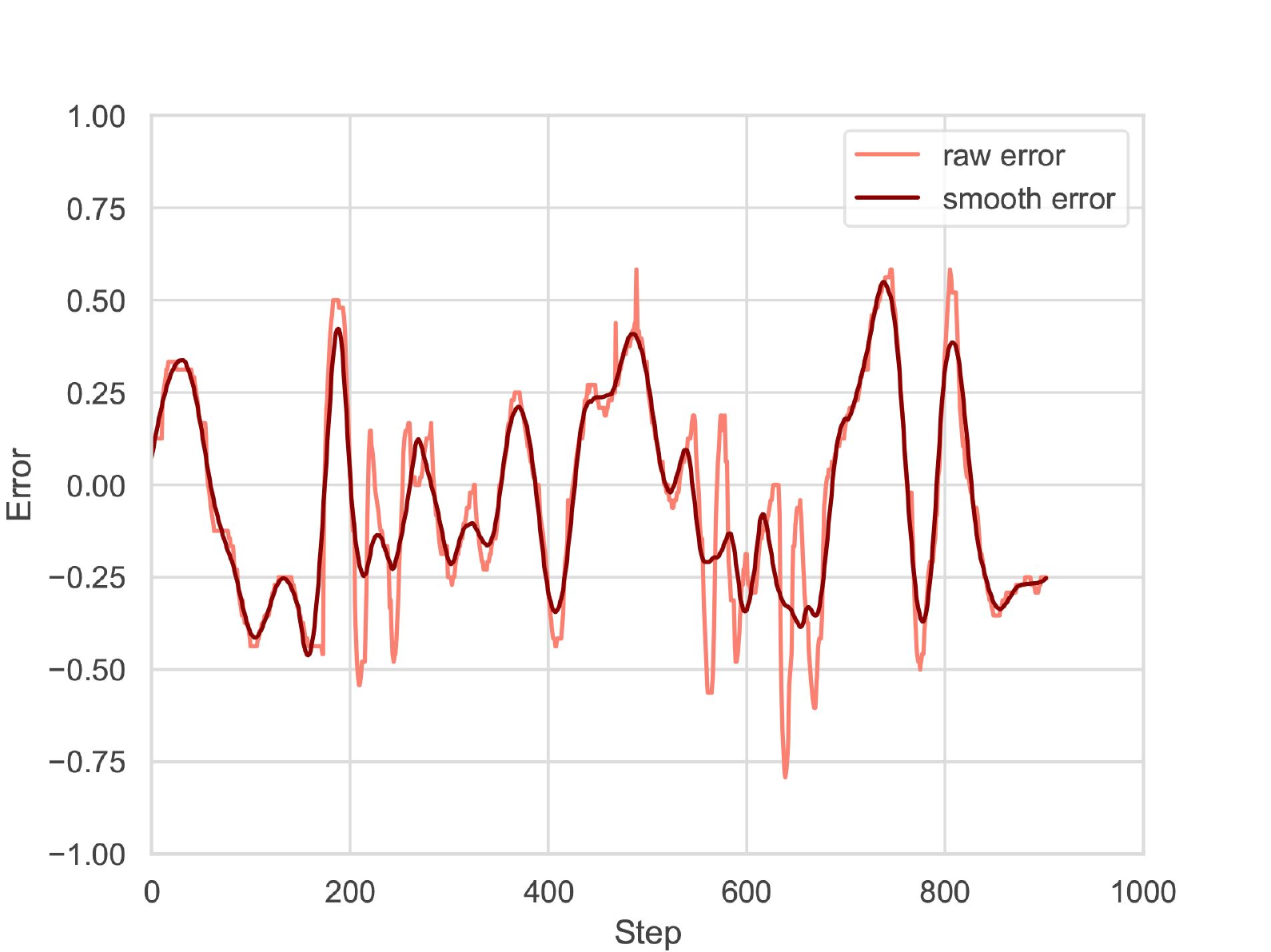}}
	
%这里是空了一行，能够实现强制将四张图分成两行两列显示，而不是放不下图了再换行，使用\\也行。
	\subfigure[Real path 2]{
		\label{map2_real}
		\includegraphics[scale=0.052]{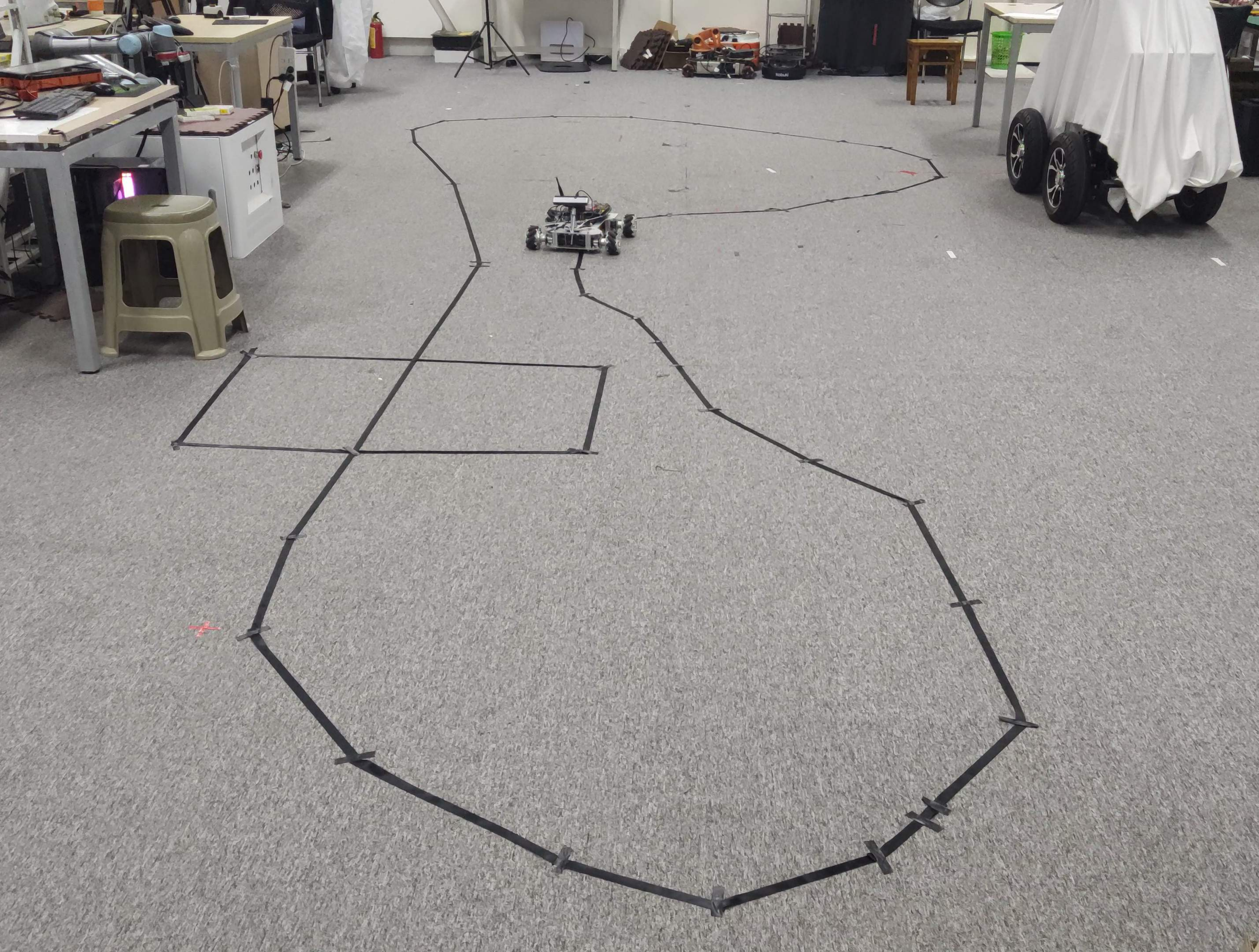}}
	\quad
    \quad
	\subfigure[The error of SAC-PID controller on Real path 2]{
		\label{map2_error_real}
		\includegraphics[height=4.5cm]{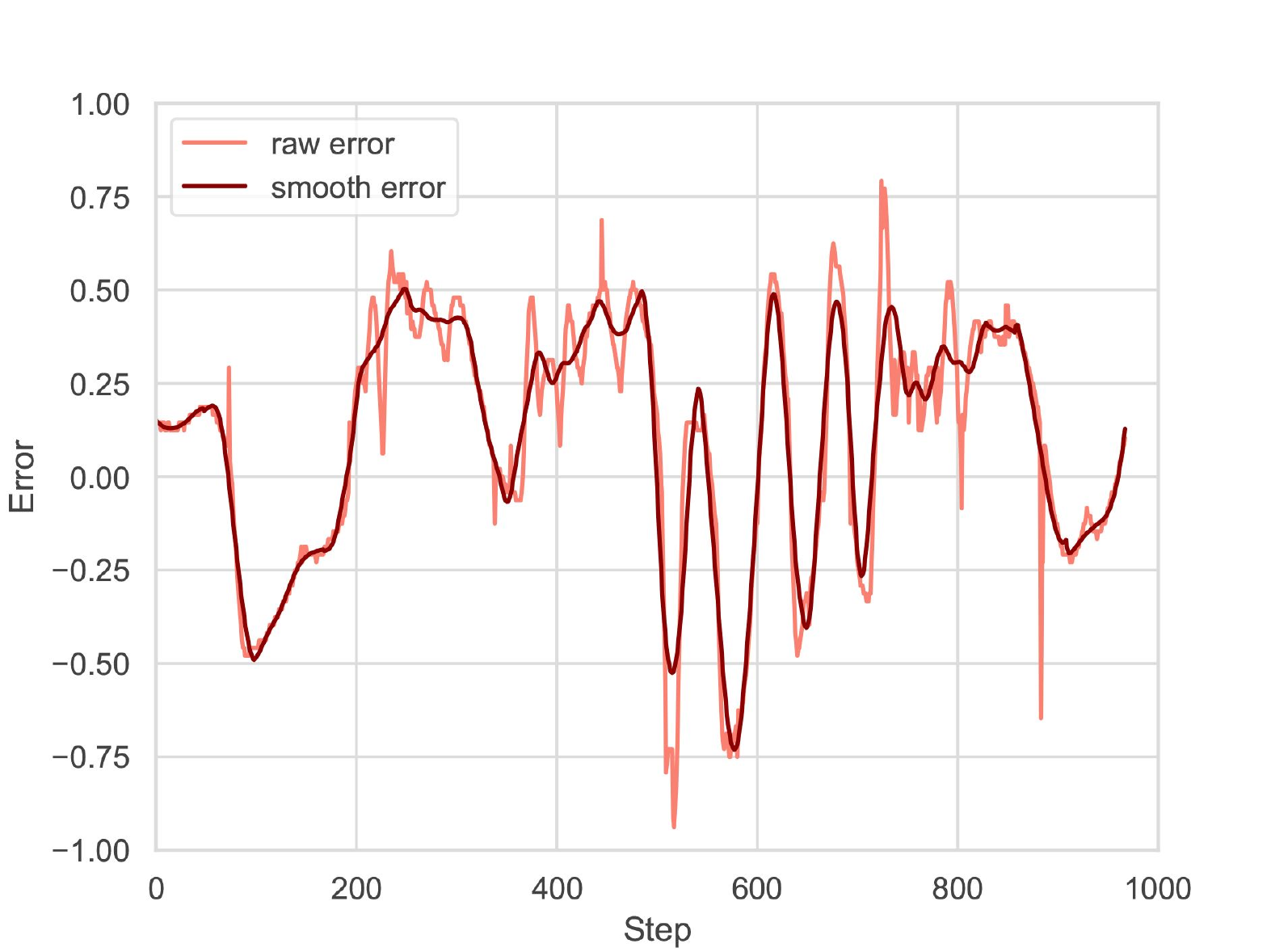}}
	\caption{The error curves on two real path}
	\label{The performance of real world}
\end{figure}

We also conducted evaluations on a Mecanum mobile robot equipped with NVIDIA Jetson TX2 and the RGB camera. We employed Model 4 trained on gazebo as the policy parameters of the SAC-PID control to test two paths of different difficulty, as shown in Fig. \ref{map1_real} and Fig. \ref{map2_real}. The quantitative evaluation results are shown in Fig. \ref{map1_error_real}, Fig. \ref{map2_error_real} and Table \ref{The success rate of each path in real world}. In the experiment we observed the SAC-PID control method has achieved good results in the real environment. The success rate of SAC-PID is as high as $80\%$ in both simple paths without forks and complex paths with forks, which represents SAC-PID control has a certain generalization from simulation to reality. Although the error compensation performance of SAC-PID control method has a certain gap between simulation and reality due to the real mobile robot has a larger time lag than simulation, the SAC-PID control is also qualified for the line-following robot on real mobile robot.
%However, due to the real mobile robot has a larger time lag than simulation, the error compensation performance of SAC-PID control method has a certain gap between simulation and reality. we leave it as a future problem to narrow down the range of error and to improve the real-time performance.

\begin{table}[htp]
\setlength{\belowcaptionskip}{0cm}
\setlength{\abovecaptionskip}{0.cm}
\centering
\caption{Evaluation results on two real paths}
\begin{tabular}{ccccc}
\toprule
\makecell[c]{Path}  & \makecell[c]{Testing number} & \makecell[c]{Success number} & \makecell[c]{Success rate(\%)}  &\makecell[c]{Average velocity(m/s)} \\
\midrule
\makecell[c]{Real path 1} & \makecell[c]{5}              & \makecell[c]{4}                                 & \makecell[c]{80}   & \makecell[c]{0.237 $\pm$ 0.017}                               \\
\makecell[c]{Real path 2} & \makecell[c]{5}             & \makecell[c]{4}                                 & \makecell[c]{80}    &  \makecell[c]{0.229 $\pm$ 0.012}                            \\
\bottomrule
\end{tabular}
\label{The success rate of each path in real world}
\end{table}

\clearpage

\section{Conclusion}\label{sec5}
In this paper, a novel model-free self-adaptive SAC-PID control approach based on reinforcement learning for automatic control of mobile robots is proposed.
The new hierarchical structure was developed, which includes the soft actor-critic layer as the upper controller and the incremental PID control layer as the lower controller.
The proposed model-free SAC-PID control architecture can adaptively drive the parameter selections of the incremental PID controllers for mobile robots without tedious tuning task. This particular point makes the SAC-PID control method robust for implementations in different systems with different variant, unknown and complex environment.
Using the presented approach for line-following robots, the parameters of incremental PID controllers can be adjusted in real time to achieve optimal control.
In this sense, the errors between the center of the mobile robot and the following line were compensated in real time. Compared with fuzzy PID control, the strong robustness, generalization and real-time performance of the SAC-PID control approach are verified through simulations of different difficulty paths on Gazebo. Furthermore, the effectiveness of SAC-PID control method is also demonstrated on a mecanum mobile Robot in real world environments.
%The test results of the four different training models in corresponding difficulty paths show that the proposed SAC-PID control approach has strong stability and real-time performance. In addition, the test results of the same training model in different paths show the good generalization of the SAC-PID control method.
In a more broad sense, since the proposed method is able to control different systems, it can also extend the SAC-PID control approach to different robotic platforms, such as robotic manipulators.

\section*{Acknowledgments}
This paper was supported by National Key Research and Development Plan Intelligent Robot Key Project(2018YFB1308402).

\clearpage
%\nocite{*}% Show all bib entries - both cited and uncited; comment this line to view only cited bib entries;
\bibliography{wileyNJD-AMA}%
\clearpage

\section{appendix\label{appendix}}
\subsection{Appendix A\label{appendix A}}
Table \ref{Hperparameter of SAC-PID} lists the hyperparameters SAC-PID employed in training phase.

\begin{table}[htp]
\caption{The Hyperparameters of SAC-PID}
\setlength{\belowcaptionskip}{0.cm}
\setlength{\abovecaptionskip}{0.cm}
\centering
\begin{tabular} {p{7cm}|p{4cm}}
\toprule
\multicolumn{1}{l|}{Parameter}         & \multicolumn{1}{l}{Value} \\
\midrule
optimizer                              & Adam                       \\
nonlinearity                           & ReLU                       \\
temperature parameter($\alpha$)                  & 0.05                       \\
learning rate($\lambda_V$, $\lambda_Q$ and $\lambda_\pi$)                        & 0.0003                     \\
discount rate($\gamma$)                          & 0.99                       \\
replay buffer size                     & $2\times10^6$                     \\
numbers of hidden layers(all networks) & 2                          \\
numbers of hidden units of per layer   & 512                        \\
numbers of minibatch(b)                & 512                        \\
target smoothing coefficient($\chi$)           & 0.005                      \\
target update interval                 & 1                          \\
gradients step                         & 1                           \\
proportional coefficient($\eta$)       & 0.5                         \\
\bottomrule
\end{tabular}
\label{Hperparameter of SAC-PID}
\end{table}

\subsection{Appendix B\label{appendix B}}
\begin{figure}[htbp]
\centerline{\includegraphics[scale=0.7]{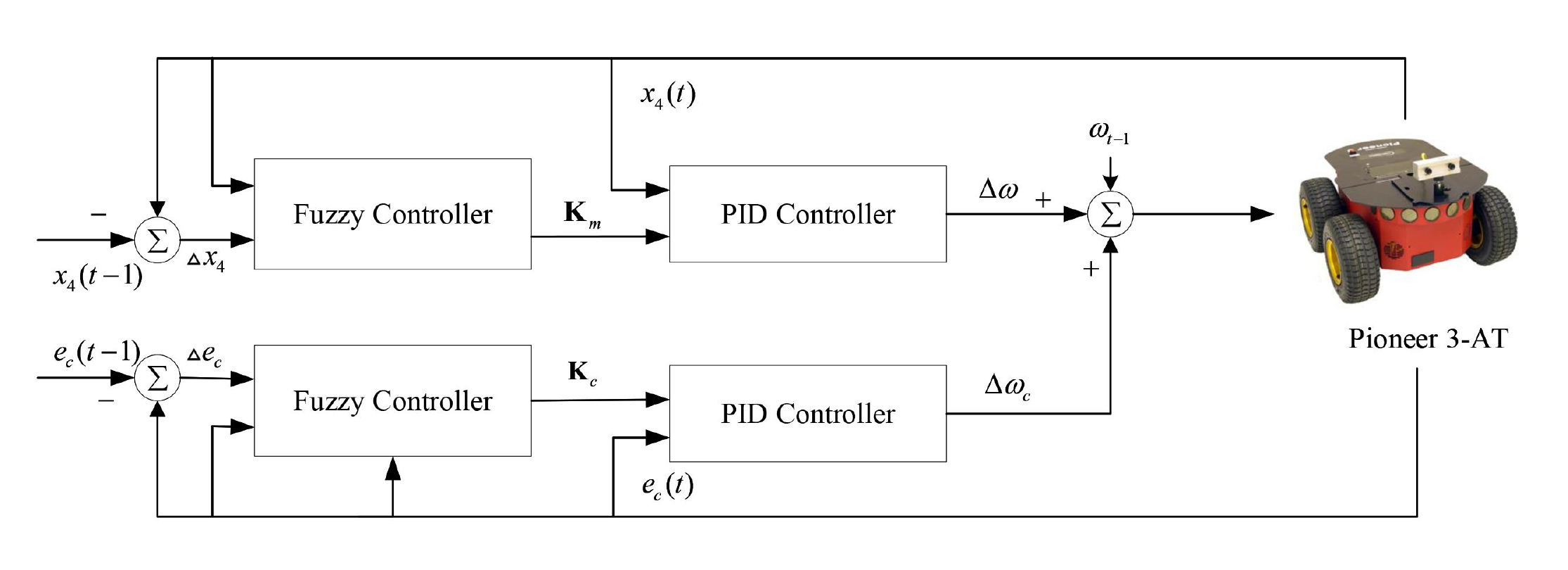}}
\caption{The structure of fuzzy PID controller.\label{fuzzy_PID_structure}}
\end{figure}

The design of the overall structure of the fuzzy PID controller is shown in the Fig. \ref{fuzzy_PID_structure}. To be specific, fuzzy PID control is composed of two fuzzy controllers with the same control rules and two incremental PID controllers with the same structure as SAC-PID controller. The inputs of the fuzzy controller are $\vartriangle {{x}_{4}}$, ${{x}_{4}}(t)$ and $\vartriangle {{e}_{c}}$, ${{e}_{c}}(t)$ and the outputs are $\vartriangle{\omega}$ and $\vartriangle {{\omega }_{c}}$, where $\vartriangle {{x}_{4}}={{x}_{4}}(t)-{{x}_{4}}(t-1)$, $\vartriangle {{e}_{c}}={{e}_{c}}(t)-{{e}_{c}}(t-1)$, ${{\mathbf{K}}_{m}}$ and ${{\mathbf{K}}_{c}}$ are the parameters of two incremental PID controllers that are adjusted by the fuzzy controllers in real time. In addition, the specific parameters of fuzzy PID are shown in Table \ref{The parameter of fuzzy PID controller}.

\begin{table}[htbp]
\caption{The parameters of fuzzy PID control}
\setlength{\belowcaptionskip}{0.cm}
\setlength{\abovecaptionskip}{0.cm}
\centering
\begin{tabular} {p{3cm}|p{3cm}|p{3cm}|p{3cm}}
\toprule
Parameter & Range        & Parameter & Range          \\
\midrule
$x_4(t)$         & {[}-1,1{]}   & $e_c(t)$         & {[}-1,1{]}     \\
$\vartriangle{x_4}$     & {[}-1,1{]}   & $\vartriangle{e_c}$      & {[}-1,1{]}     \\
$k_{mp}$     & {[}-20,20{]} & $k_{cp}$      & {[}-0.1,0.1{]} \\
$k_{mi}$       & {[}0,0.5{]}  & $k_{ci}$       & {[}-0.1,0.1{]} \\
$k_{md}$       & {[}0,0.1{]}  & $k_{cd}$         & {[}0,0.1{]} \\
\bottomrule
\end{tabular}
\label{The parameter of fuzzy PID controller}
\end{table}

\end{document}